\documentclass[]{xiaomiev}

\usepackage[toc,page,header]{appendix}
\usepackage{minitoc}
\usepackage{solarized-light}
\usepackage{booktabs}
\usepackage{multirow}
\usepackage{graphicx}
\usepackage{array}
\usepackage{siunitx,array}
\usepackage[table]{xcolor}
\usepackage{makecell,array,booktabs}
\usepackage{makecell}
\usepackage{framed}

\usepackage{bbding}
\usepackage{hyperref}
\usepackage{amsmath} 
\usepackage{placeins}
\usepackage[colorinlistoftodos]{todonotes}
\usepackage{longtable}
\usepackage{hhline}
\usepackage{fancyvrb}
\usepackage{fvextra}
\usepackage{CJKutf8}
\usepackage{multicol}
\usepackage{cleveref}
\usepackage{tablefootnote}
\usepackage{threeparttable}
\usepackage{tabularx}
\usepackage{mdframed}
\usepackage{subcaption}
\usepackage[usestackEOL]{stackengine}
\usepackage[numbers]{natbib}
\newcommand{\commentout}[1]{}
\renewcommand{\paragraph}[1]{\noindent\textbf{#1.}}

\usepackage{enumitem}
\usepackage{hyperref}

\setlist[itemize]{leftmargin=15pt}
\makeatletter
\providecommand{\@bottomtitlebar}{}
\makeatother

\RequirePackage{xspace}
\makeatletter
\DeclareRobustCommand\onedot{\futurelet\@let@token\@onedot}
\def\@onedot{\ifx\@let@token.\else.\null\fi\xspace}

\makeatother
\newcommand{\Ours}{\texttt{Xiaomi-Robotics-0}}

\title{\mbox{Xiaomi-Robotics-0: An Open-Sourced Vision-Language-Action}\\ Model with Real-Time Execution
}
\author[\textcolor{xiaomievblue}{1}]{Xiaomi Robotics}

\abstract{
In this report, we introduce \texttt{\Ours}, an advanced vision-language-action (VLA) model optimized for high performance and fast and smooth real-time execution.
The key to our method lies in a carefully designed training recipe and deployment strategy.
\texttt{\Ours} is first pre-trained on large-scale cross-embodiment robot trajectories and vision-language data, endowing it with broad and generalizable action-generation capabilities while avoiding catastrophic forgetting of the visual-semantic knowledge of the underlying pre-trained VLM.
During post-training, we propose several techniques for training the VLA model for asynchronous execution to address the inference latency during real-robot rollouts.
During deployment, we carefully align the timesteps of consecutive predicted action chunks to ensure continuous and seamless real-time rollouts.
We evaluate \texttt{\Ours} extensively in simulation benchmarks and on two challenging real-robot tasks that require precise and dexterous bimanual manipulation.
Results show that our method achieves state-of-the-art performance across all simulation benchmarks.
Moreover, \texttt{\Ours} can roll out fast and smoothly on real robots using a consumer-grade GPU, achieving high success rates and throughput on both real-robot tasks.
To facilitate future research, code and model checkpoints are open-sourced at \url{https://xiaomi-robotics-0.github.io}
}

\begin{document}
\maketitle
\footnotetext[1]{See Contributions section for full author list. Please send correspondence to \href{mailto:mi-robotics@xiaomi.com}{\mbox{mi-robotics@xiaomi.com}}.}

\begin{figure*}[!t]
\centering
\includegraphics[width=0.95\textwidth]{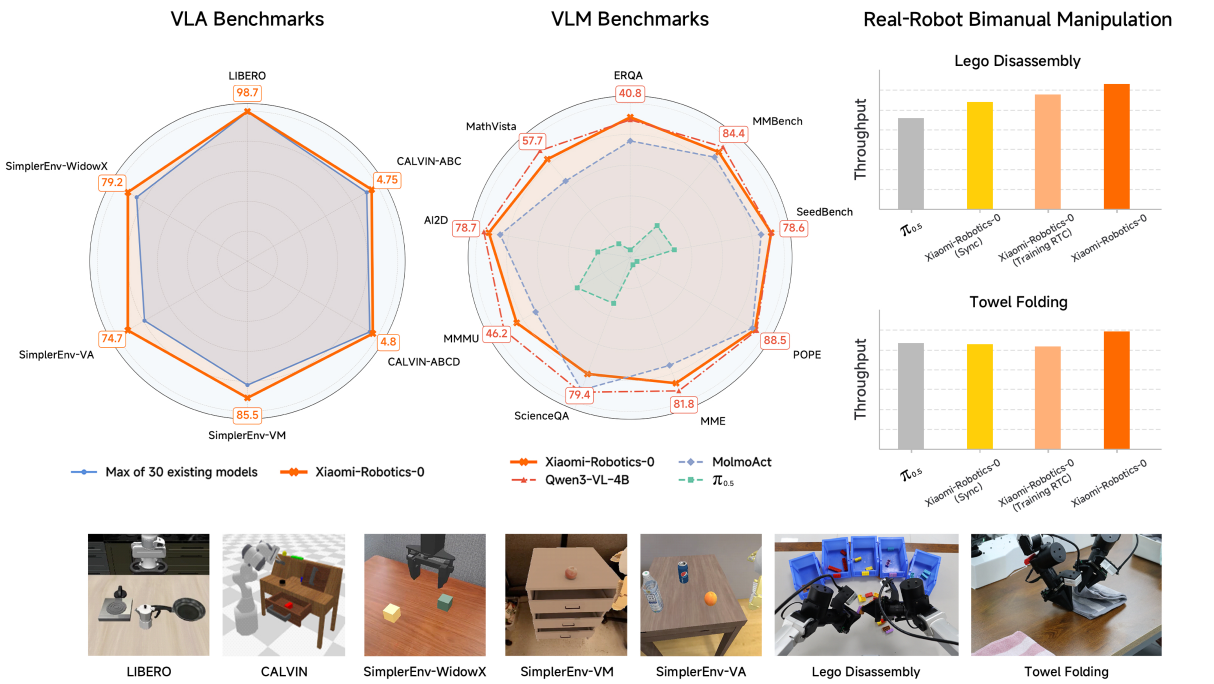}
\caption{
\textbf{Overview.} 
\Ours{} achieves state-of-the-art performance in three widely-used simulation benchmarks.
It also attains high throughput on two challenging real-robot bimanual manipulation tasks.
Furthermore, it matches the underlying pre-trained VLM on several VLM benchmarks.
}
\label{fig:overview}
\end{figure*}

\section{Introduction}
Vision-language-action (VLA) models have emerged as a new paradigm for effective robot policy learning~\cite{zitkovich2023rt2, kim2024openvla, black2024pi_0}.
Building upon pre-trained vision-language models (VLMs), VLA models provide a unified framework that maps observations and language instructions directly to actions across a wide range of tasks.
However, despite their strong performance and generalization capabilities, VLA models suffer from high inference latency due to their large parameter counts, which can scale to billions of parameters.
This creates challenges for smoothly chaining actions across consecutive inference steps, leading to out-of-distribution jerky motions if not handled properly~\cite{black2025real}.

In this report, we introduce \Ours{} (Fig.~\ref{fig:overview}), an advanced vision-language-action (VLA) model that delivers high performance while enabling fast and smooth rollouts on real robots.
It is composed of a pre-trained vision-language model (VLM)~\cite{bai2025qwen3} for processing vision and language inputs and a diffusion transformer~\cite{peebles2023scalable} for generating actions via flow-matching~\cite{liu2022flow, lipman2022flow}.
The training recipe contains two stages: pre-training and post-training.
During pre-training, we train the model with large-scale cross-embodiment robot trajectories and vision-language data.
This stage endows the model with broad and generalizable action generation capabilities while maintaining the strong vision-language capabilities of the underlying pre-trained VLM it built upon.
During post-training, we introduce novel techniques to enable fast and smooth asynchronous execution during real-robot rollouts.
Specifically, we first condition the generation of action chunks by prefixing it with actions from the previous inference as in~\cite{black2025training}.
While this conditioning method ensures continuity across consecutively generated chunks, it allows the generation of later-timestep actions to exploit the temporal correlation that successive actions tend to be similar.
As a result, policy learning can take a shortcut by simply imitating the action prefix rather than attending to visual and language signals, resulting in less reactive policies and degraded performance.
To address this issue, we replace the causal attention mask with a $\Lambda$-shape attention mask~\cite{jiang2024minference, xiao2023efficient, han2024lm} during post-training, encouraging action generation to pay more attention to visual and language conditions rather than over-relying on the action prefix.
During deployment, we carefully align the timesteps of action chunks generated from consecutive inferences to ensure continuous and seamless real-robot rollouts.

We evaluate \Ours{} extensively on both simulation benchmarks and a bimanual real-robot platform.
Our model achieves state-of-the-art performance across three widely-used simulation benchmarks.
Specifically, it achieves an average success rate of \textbf{98.7\%} on \textbf{LIBERO}~\cite{liu2023libero}.
On \textbf{SimplerEnv}~\cite{li2024evaluating}, it delivers strong performance under the visual matching (\textbf{85.5\%}) and visual aggregation (\textbf{74.7\%}) settings in the Google Robot evaluations as well as the WidowX evaluations (\textbf{79.2\%}).
On \textbf{CALVIN}~\cite{mees2022calvin}, \Ours{} improves the average length of completing 5 tasks in a row from 4.54 to \textbf{4.75} and from 4.67 to \textbf{4.80} on the ABC$\rightarrow$D and ABCD$\rightarrow$D split, respectively.
In real-robot experiments, we evaluate on two challenging tasks that require precise and dexterous bimanual manipulation: Lego Disassembly and Towel Folding.
\Ours{} is able to achieve high success rates and outperforms state-of-the-art methods~\cite{intelligence2025pi_0_5, black2025training} on both tasks in terms of throughput, enabling smooth real-time execution.
In addition, our pre-trained model matches the performance of the underlying pre-trained VLM~\cite{bai2025qwen3} on several general vision-language benchmarks and a benchmark focused on embodied reasoning~\cite{team2025gemini}.

We release the pre-trained and post-trained checkpoints, along with the inference code to facilitate future research.
We hope these resources serve as a practical foundation for advancing vision-language-action (VLA) models.

\section{Xiaomi-Robotics-0}
\label{methods}
\Ours{} is an end-to-end vision-language-action (VLA) model that takes as inputs observation images, a language instruction, and the robot proprioceptive state.
It outputs an action chunk~\cite{zhao2023learning} to control a bimanual robot in an end-to-end manner.

\begin{figure}[t]
    \centering
    \includegraphics[width=\linewidth]{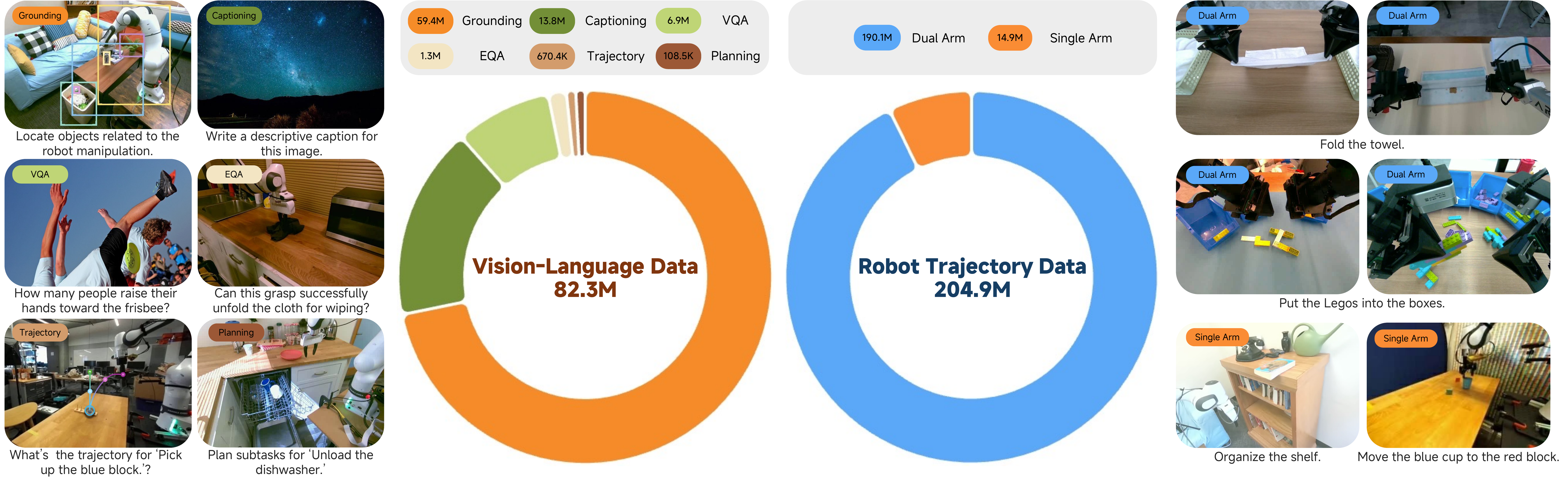}
    \caption{
    \textbf{Data.}
    \Ours{} leverages both robot trajectory data and vision-language (VL) data during pre-training.
    }
    \label{fig:data_composition}
\end{figure}

\subsection{Data}
\label{methods:data}
We leverage both robot trajectory data and vision-language (VL) data during training.
Fig.~\ref{fig:data_composition} illustrates the detailed data composition.
Our robot trajectory data are sourced from multiple open-sourced robot datasets (\textit{e.g.}, DROID~\cite{khazatsky2024droid} and MolmoAct~\cite{lee2025molmoact}) as well as in-house data collected by ourselves.
Our in-house data consists of teleoperated trajectories for two challenging tasks: Lego Disassembly and Towel Folding.
In total, we collected 338 and 400 hours of data for these two tasks, respectively.
Overall, the entire robot trajectory dataset contains about 200M timesteps for training.

For the vision-language data, we curate a comprehensive corpus of more than 80M samples from two primary sources: general vision-language (VL) datasets~\cite{sharma2018conceptual, changpinyo2021cc12m, tong2024cambrian1fullyopenvisioncentric, wiedmann2025finevisionopendataneed} and robot datasets~\cite{lee2025molmoact, khazatsky2024droid}.
While general VL data preserve broad semantic knowledge, VL data derived from robot trajectories enhance the model's perception on robot-centric images, which are often captured from egocentric perspectives or wrist-mounted cameras.
Specifically, we curate data by focusing on four vision-language tasks: (1) visual grounding, (2) visual question answering (VQA), (3) image captioning, and (4) embodied reasoning \& planning.
For visual grounding, we develop a rigorous cross-validated consensus mechanism that integrates Grounded SAM~\cite{ren2024grounded}, Grounding DINO 1.5~\cite{ren2024grounding}, and LLMDet~\cite{fu2025llmdet}, ensuring pixel-level annotation precision.
VQA and captioning annotation quality is further refined through re-labeling using state-of-the-art pre-trained VLMs~\cite{bai2025qwen3}.
For embodied reasoning \& planning, we leverage pre-trained VLMs to generate data from root trajectories, focusing on embodied question answering (EQA), high-level task planning, and point trajectory prediction.

\begin{figure}[t]
    \centering
    \includegraphics[width=\linewidth]{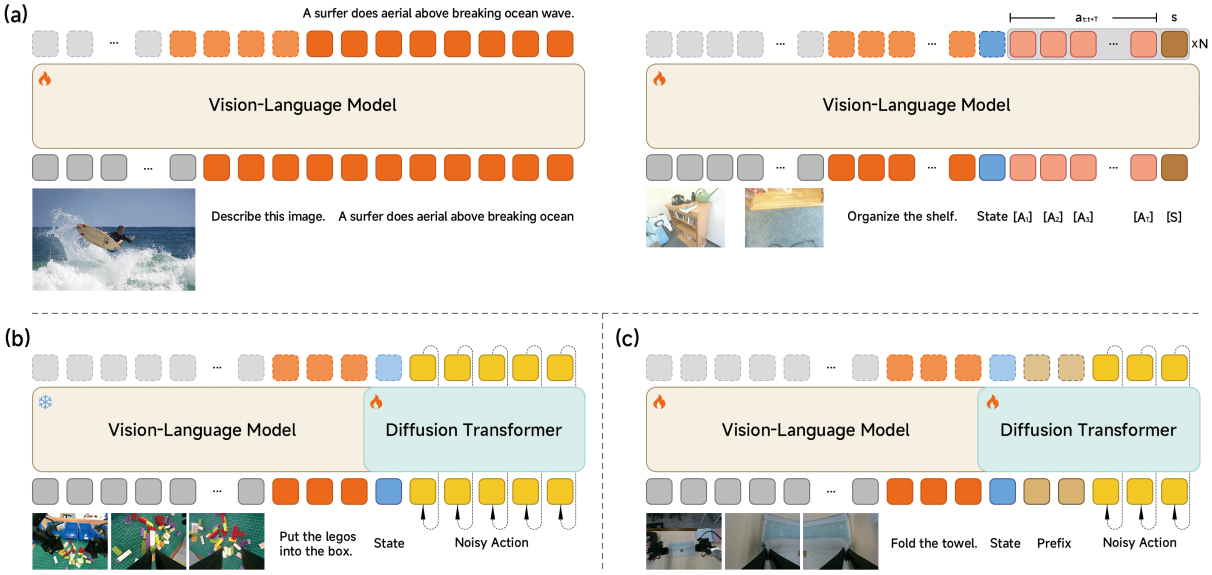}
    \caption{
        \textbf{Model \& Training.}
        (a) During the first step of pre-training, we train the VLM on both vision-language data (left) and robot trajectory data (right).
        Vision-language data are trained via a next-token-prediction objective.
        We adopt the training paradigm in Choice Policies~\cite{qi2025coordinated} to train the VLM for action prediction on the robot trajectory data.
        (b) In the second step of pre-training, we freeze the VLM and train the diffusion transformer for generating actions via flow-matching.
        (c) During post-training for asynchrnnous execution, we prepend clean action prefix to the noisy action tokens.
    }
    \label{fig:training}
\end{figure}

\subsection{Model \& Training}
\label{methods:model_training}
\Ours{} adopts a mixture-of-transformers (MoT)~\cite{liang2024mixture} model architecture.
It consists of a pre-trained vision-language model (VLM) (\textit{i.e.}, Qwen3-VL-4B-Instruct~\cite{bai2025qwen3}) and a diffusion transformer (DiT)~\cite{peebles2023scalable}.
The VLM takes as inputs observation images $\mathbf{o}_{t}$ of the current timestep, along with a language instruction $l$ provided by the user.
The DiT generates a $T$-step action chunk $\mathbf{a}_{t:t+T}$~\cite{zhao2023learning} via flow-matching, conditioned on the KV cache produced by the VLM and the robot proprioceptive state.
In total, the model has 4.7B parameters.

\subsubsection{Pre-training}
\label{methods:model_training:pre_training}
We perform pre-training in two steps.
In the first step, our goal is to endow the VLM with action-generation capability by training it to predict action chunks from observation images, language instructions, and robot proprioceptive states.
To account for the multi-modality in trajectories, we adopt Choice Policies~\cite{qi2025coordinated} for action prediction.
Specifically, we train the VLM to simultaneously predict $N$ action chunk candidates along with their corresponding scores (right of Fig.~\ref{fig:training}(a)).
During training, we compute the $L_1$ distance between each predicted action chunk candidate and the ground truth, and utilize these values as supervision targets for score prediction.
Action prediction is supervised via a winner-takes-all scheme: only the candidate with the lowest $L_1$ distance is updated via backpropagation.

Architecturally, we encode the robot proprioceptive state $\mathbf{s}_{t}$ using an MLP.
For action prediction, we append $T$ learnable tokens $\texttt{[$A_i$]}$ to predict $N$ sets of $T$-step action chunks $\mathbf{a}_{t:t+T}$, and one additional token $\texttt{[$S$]}$ to predict the score $s$ for each chunk.
The input token sequence is: $\mathbf{o}_{t}, l, \mathbf{s}_{t}, \texttt{[$A_1$]}, ..., \texttt{[$A_{T}$]}, \texttt{[$S$]}$.
The output of each action token $\texttt{[$A_i$]}$ is mapped to $N$ predictions of the action at the $i$-th timestep, while the output of the score token $\texttt{[$S$]}$ is mapped to $N$ scores.

To avoid catastrophic forgetting of the strong vision-language capabilities of the underlying pre-trained VLM~\cite{bai2025qwen3}, and to improve its visual understanding on robot-centric data, we co-train the model with the entire robot trajectory data \textit{and} vision-language data described in Sec.~\ref{methods:data}.
The vision-language data is trained with a next-token-prediction objective (left of Fig.~\ref{fig:training}(a)).
We sample vision-language data and robot trajectory data at a ratio of 1:6.

After the first step of training, the VLM is equipped with the capability to generate actions.
In the second step, we freeze the VLM and train the diffusion transformer (DiT) from scratch on the entire robot trajectory data with a flow-matching loss (Fig.~\ref{fig:training}(b)):
\begin{equation}
    L(\theta) = ||\mathbf{v}_{\theta}(\mathbf{o}_{t}, l, \mathbf{s}_{t}, \tilde{\mathbf{a}}_{t:t+T}^{\tau}, \tau) - \mathbf{u}(\tilde{\mathbf{a}}_{t:t+T}^{\tau}, \mathbf{a}_{t:t+T}, \tau)||^{2}_{2}
\end{equation}
$\tau \in [0, 0.999]$ is the flow-matching timestep.
$\tilde{\mathbf{a}}_{t:t+T}^{\tau} = \tau \mathbf{a}_{t:t+T} + (1 - \tau) \boldsymbol{\epsilon}$ is the noisy action where $\boldsymbol{\epsilon} \sim \mathcal{N}(\mathbf{0}, \mathbf{I})$.
Following~\cite{black2024pi_0, intelligence2025pi_0_5}, we sample $\tau$ from a Beta distribution, placing more weight on noisier timesteps during training.
We leverage adaptive normalization layers (adaLN)~\cite{perez2018film, peebles2023scalable} to inject the flow-matching timestep condition into the DiT for action generation.
The robot proprioceptive state and noisy actions are encoded with MLPs.
We add a learnable attention sink token at the front of the state and noisy action tokens to stabilize the attention distribution during training.
The input tokens for the DiT are sequenced as: $\texttt{[SINK]}, \mathbf{s}_{t}, \tilde{\mathbf{a}}_{t}, \ldots, \tilde{\mathbf{a}}_{t+T-1}$, where $\tilde{\mathbf{a}}_{t+i}$ corresponds to a noisy action.
To account for the temporal relationship between actions at different timesteps, we use causal attention in the DiT.
To reduce inference latency, we use a 16-layer DiT and condition it on the KV cache from the last 16 layers of the VLM.
During this step, we leverage the VLM as a frozen multimodal conditioner for providing visual-language features, while the DiT learns to generate action chunks conditioned on these features.
The inputs to the VLM contains only the observation images $\mathbf{o}_{t}$ and language $l$ without the newly introduced tokens for action prediction as in the first step.

\begin{figure}
    \centering
    \includegraphics[width=0.5\linewidth]{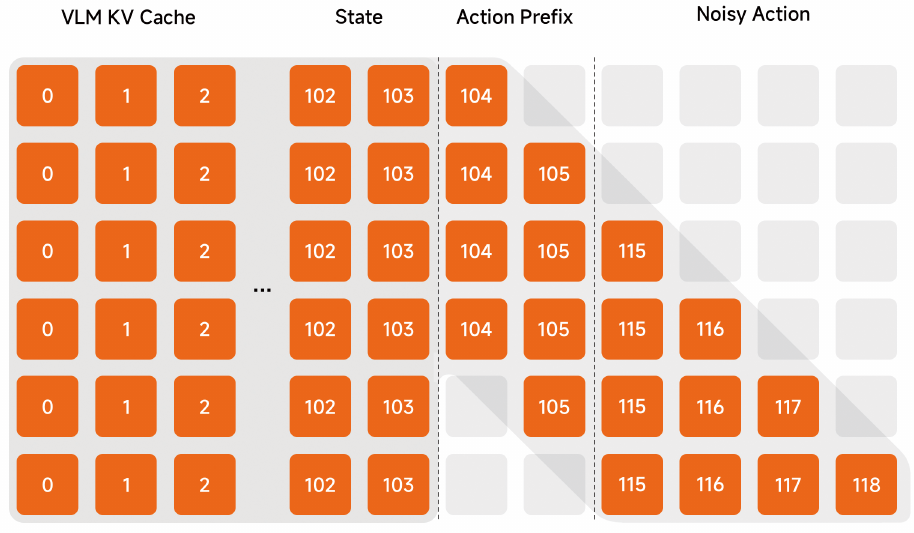}
    \caption{
       \textbf{The $\Lambda$-Shape Attention Mask for Post-Training}.
       A noisy action token can only attend to the vision and language tokens via the VLM KV cache, the sink token, the state token, and the action tokens of the previous $w$ timesteps.
       The number in each token indicates the RoPE positional index of the token.
       Note that we add an offset of 10 to the positional indices of the noisy action tokens to allow the model to distinguish them from the clean action prefix tokens.
    }
    \label{fig:attention}
\end{figure}

\subsubsection{Post-training}
\label{methods:model_training:post_training}
In post-training, we adapt \Ours{} to a specific robot by training solely on the trajectory data of the robot.
We describe two post-training methods for synchronous and asynchronous execution, respectively.
For synchronous execution, we simply unfreeze the entire model, \textit{i.e.}, the VLM and DiT, and continue training on predicting actions via flow-matching as in the second step of pre-training.

However, when deploying on real robots, inference latency becomes non-negligible due to the large number of parameters, causing pauses in synchronous execution where the robot remains idle until the next inference is completed.
Asynchronous execution enables the robot to continue rolling out trajectories during model inference.
In this setting, it is crucial to maintain consistency across consecutively inferred action chunks and ensure smooth transitions between them, since inconsistency can induce jerky motions and drive the robot into out-of-distribution regimes~\cite{black2025real}.
To address this problem, prior work proposes real-time chunking (RTC)~\cite{black2025real} and training-time RTC~\cite{black2025training}, which condition action generation on previously committed actions.
In this work, we follow training RTC~\cite{black2025training} and condition action generation on $\Delta t_{c}$ previously committed actions by prefixing them to the noisy action tokens in DiT (Fig.~\ref{fig:training}(c)).
The input token sequence of the DiT thus becomes: $\texttt{[SINK]}, \mathbf{s}_{t}, \mathbf{a}_{t}, \ldots, \mathbf{a}_{t+\Delta t_{c}-1}, \tilde{\mathbf{a}}_{t+\Delta t_{c}}^{\tau}, \ldots, \tilde{\mathbf{a}}_{t+T-1}^{\tau}$.
While this approach reduces inconsistency, it also enables predictions of later-timestep actions to exploit the temporal correlation between successive actions.
As a result, policy learning may take a shortcut by simply copying the action prefix instead of attending to the visual and language inputs, leading to less reactive policies and degraded performance.

We propose simple techniques to alleviate this issue.
We first simply add an offset to the RoPE positional indices of the noisy action tokens to enable the model to distinguish tokens of noisy actions from those of the clean action prefix.
In addition, we change the original causal attention mask of the DiT to a $\Lambda$-shape attention mask~\cite{jiang2024minference, xiao2023efficient, han2024lm} (Fig.~\ref{fig:attention}).
Since the noisy action tokens immediately following the tokens of the action prefix can attend to them, the generated actions can smoothly transition from the action chunk produced by the previous inference.
In contrast, noisy action tokens of later timesteps cannot attend to the tokens corresponding to the conditioned action prefix, forcing them to attend to other signals (\textit{e.g.}, visual observations and languages), thus ensuring reactivity in the predicted actions.

During training, we sample $\Delta t_{c}$ from the set $\{0, 1, \cdots , 6\}$.
When $\Delta t_{c} > 0$, we dynamically re-weight the flow-matching loss based on the $L_1$ error between the \textit{online-predicted} actions and the ground-truth actions.
This strategy prioritizes samples with larger deviations, directing the model to focus on correcting significant execution errors.

\begin{figure}
    \centering
    \includegraphics[width=0.65\linewidth]{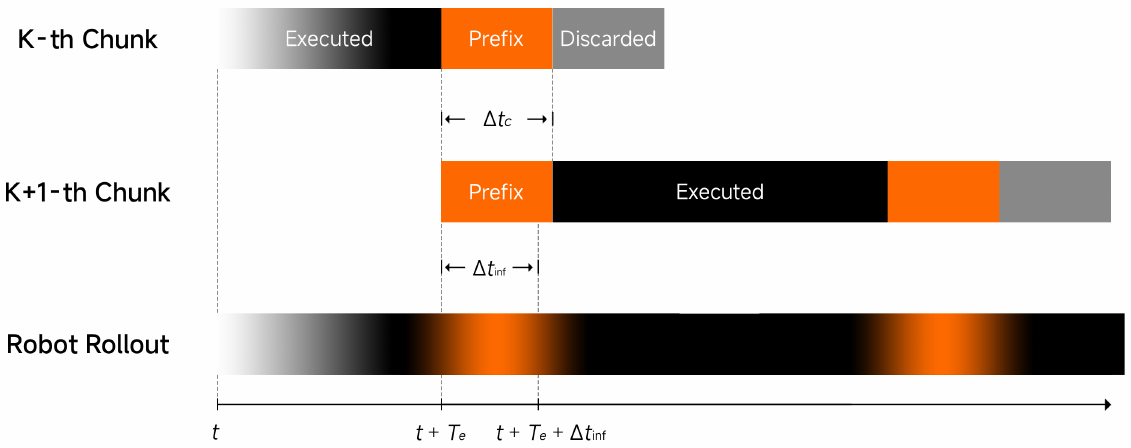}
    \caption{
        \textbf{Asynchronous Execution.}
        We show two consecutive chunks and how they are stitched together during robot rollout.
        See Sec.~\ref{methods:training:deployment} for more details.
    }
    \label{fig:deployment}
\end{figure}

\subsection{Deployment}
\label{methods:training:deployment}
We describe deployment methods for synchronous and asynchronous execution, respectively.

\paragraph{Synchronous Execution}
For synchronous execution, we control the robot to execute the first $T_{e}$ steps of actions within the $T$-step predicted action chunk.
Once these actions have been rolled out, we immediately start inferring the next action chunk using the latest observation images and proprioceptive state.
The robot remains idle until the inference completes.

\paragraph{Asynchronous Execution}
We visualize the asynchronous execution in Fig.~\ref{fig:deployment}.
For each inferred action chunk, we first similarly roll out $T_{e}$ steps before triggering the subsequent inference cycle.
However, instead of staying idle, the robot continues executing the remaining actions of the current chunk while the next chunk is being inferred.
We condition the next inference by prefixing the noisy actions with the actions from step $T_e$ to step $T_e + \Delta t_{c} - 1$ of the current chunk.
Upon completion, the newly generated chunk is executed starting from step $\Delta t_{\mathrm{inf}}$, where $\Delta t_{\mathrm{inf}}$ is the inference latency.
We set $\Delta t_{c} \ge \Delta t_{\mathrm{inf}}$ so that the action prefix covers the entire inference window.
As a result, there are always actions available for execution throughout the entire inference, enabling seamless transition between consecutive inference cycles.

During inference, we initialize the action chunk by sampling from a standard Gaussian distribution, $\mathbf{a}^{\tau=0}_{t:t+T} \sim \mathcal{N}(\mathbf{0}, \mathbf{I})$.
We then perform 5 flow-matching steps and integrate $\tau$ from 0 to 1 to obtain the predicted action chunk.
Deployed on an NVIDIA GeForce RTX 4090 GPU, the model achieves an inference latency of $t_{\mathrm{inf}}=80\rm{ms}$.
To ensure consistency with the training distribution, we synchronize all input modalities by resampling them to a unified 30Hz timeline using timestamps.
At each clock tick, the temporally nearest measurements from all sensors are aggregated to form a synchronized model input.

\section{Experiments}
\subsection{Simulation Benchmarks}
\label{experiments:simulation_benchmarks}
We evaluate our method on three widely-used simulation benchmarks.
\begin{itemize}
  \item \textbf{LIBERO}~\cite{liu2023libero}:
  LIBERO features a robot arm performing various manipulation in the simulation.
  We use the filtered expert demonstrations from~\cite{kim2024openvla}, which remove unsuccessful trajectories, for training.
  Following~\cite{pertsch2025fast}, we train the model on data from all four splits, \textit{i.e.}, Libero-Spatial, Libero-Object, Libero-Goal, and Libero-Long.
  We follow the standard evaluation protocol in OpenVLA~\cite{kim2024openvla} and report success rates on each split and the average success rate across all splits.
  We set the action chunk length $T=10$.
  
  \item \textbf{CALVIN}~\cite{mees2022calvin}:
  CALVIN contains four different environments (\textit{i.e.}, A, B, C, and D) in total.
  It is designed for multi-task learning and long-horizon manipulation.
  We follow the standard evaluation protocol~\cite{mees2022calvin} and evaluate on the ABCD$\rightarrow$D and ABC$\rightarrow$D splits.
  For ABCD$\rightarrow$D, the model is trained on data collected from environments A, B, C, and D; for ABC$\rightarrow$D, it is trained on data collected from environments A, B, and C only.
  In both settings, evaluation is conducted in environment D.
  Therefore, ABCD$\rightarrow$D measures in-distribution performance, whereas ABC$\rightarrow$D quantifies out-of-distribution generalization capabilities.
  During evaluation, the model is prompted with 1000 unique instruction chains, each containing five instructions.
  For each chain, the model outputs actions to control the robot to solve tasks specified by the instructions sequentially.
  We report success rates of completing 1, 2, 3, 4, and 5 tasks in a row and the average length of tasks completed per chain.
  We set the action chunk length $T=10$.
  
  \item \textbf{SimplerEnv}~\cite{li2024evaluating}:
  SimplerEnv is a real-to-sim benchmark featuring two robot platforms: Google Robot and WidowX.
  It enables a pipeline where policies are trained on real-world robot trajectories and subsequently evaluated in simulated environments.
  The Google Robot environment provides two evaluation settings: Visual Matching and Variant Aggregation.
  Visual Matching aligns the visual appearance between real-robot scenes and their simulated counterparts, whereas Variant Aggregation introduces visual randomization to test robustness.
  Following the standard evaluation protocol of SimplerEnv~\cite{li2024evaluating}, we train our policy on the RT-1 Fractal dataset~\cite{brohan2022rt} and evaluate it on four tasks in the Google Robot environment under both settings.
  For the WidowX environment, we train our policy on the Bridge dataset~\cite{walke2023bridgedata}.
  For each policy, we report success rates for all tasks and variants, as well as the average success rate across tasks.
  We set the action chunk length $T=4$.
\end{itemize}

\begin{table*}[t]
\centering
\begin{tabular}{lccccc}
\toprule
\textbf{Method}   & \textbf{Libero-Spatial} & \textbf{Libero-Object} & \textbf{Libero-Goal} & \textbf{Libero-Long} & \textbf{Average}  \\ \midrule
OpenVLA~\cite{kim2024openvla}                   & 84.7$\%$ & 88.4$\%$ & 79.2$\%$ & 53.7$\%$ & 76.5$\%$ \\
OpenVLA-OFT~\cite{kim2025fine}                  & 97.6$\%$ & 98.4$\%$ & 97.9$\%$ & 94.5$\%$ & 97.1$\%$ \\
$\pi_{0}$~\cite{black2024pi_0}                  & 96.8$\%$ & 98.8$\%$ & 95.8$\%$ & 85.2$\%$ & 94.2$\%$ \\
$\pi_{0}$-FAST~\cite{pertsch2025fast}           & 96.4$\%$ & 96.8$\%$ & 88.6$\%$ & 60.2$\%$ & 85.5$\%$ \\
$\pi_{0.5}$~\cite{intelligence2025pi_0_5}       & 98.8$\%$ & 98.2$\%$ & 98.0$\%$ & 92.4$\%$ & 96.9$\%$ \\
GR00T-N1~\cite{gr00tn1_2025}                    & 94.4$\%$ & 97.6$\%$ & 93.0$\%$ & 90.6$\%$ & 93.9$\%$ \\
UniVLA~\cite{wang2025unified}                      & 95.4$\%$ & 98.8$\%$ & 93.6$\%$ & 94.0$\%$ & 95.5$\%$ \\
Discrete Diffusion VLA~\cite{liang2025discrete} & 97.2$\%$ & 98.6$\%$ & 97.4$\%$ & 92.0$\%$ & 96.3$\%$ \\
MemoryVLA~\cite{shi2025memoryvla}               & 98.4$\%$ & 98.4$\%$ & 96.4$\%$ & 93.4$\%$ & 96.7$\%$ \\
FLOWER~\cite{reuss2025flower}                   & 97.5$\%$ & 99.1$\%$ & 96.1$\%$ & 94.9$\%$ & 96.9$\%$ \\
EO-1~\cite{qu2025eo}                            & \textbf{99.7}$\%$ & \underline{99.8}$\%$ & \textbf{99.2}$\%$ & \underline{94.8}$\%$ & \underline{98.2}$\%$ \\
\rowcolor[HTML]{FFE0CD}
\texttt{\Ours} (Ours) & \underline{98.8$\%$} & \textbf{100.0$\%$} & \underline{98.8$\%$} & \textbf{97.2$\%$} & \textbf{98.7$\%$} \\
\bottomrule
\end{tabular}
\caption{Results on the LIBERO benchmark.}
\label{tab:libero}
\end{table*}

\begin{table}[h]
\centering
\begin{tabular}{lccccccc}
\toprule
 & & \multicolumn{5}{c}{\textbf{Tasks Completed in a Row}} & \\ \cmidrule(lr){3-7}
\textbf{Method} & \textbf{Setting} & 1 & 2 & 3 & 4 & 5 & \textbf{Avg. Len.} $\uparrow$ \\ \midrule
RoboFlamingo~\cite{li2023vision}      & ABCD$\rightarrow$D & 96.4$\%$ & 89.6$\%$ & 82.4$\%$ & 74.0$\%$ & 66.0$\%$ & 4.09 \\
GR-1~\cite{wu2023unleashing}          & ABCD$\rightarrow$D & 94.9$\%$ & 89.6$\%$ & 84.4$\%$ & 78.9$\%$ & 73.1$\%$ & 4.21 \\
MoDE~\cite{reuss2024efficient}        & ABCD$\rightarrow$D & 97.1$\%$ & 92.5$\%$ & 87.9$\%$ & 83.5$\%$ & 77.9$\%$ & 4.39 \\
RoboVLMs~\cite{liu2025towards}        & ABCD$\rightarrow$D & 96.7$\%$ & 93.0$\%$ & 89.9$\%$ & 86.5$\%$ & 82.6$\%$ & 4.49 \\
MDT~\cite{reuss2024multimodal}        & ABCD$\rightarrow$D & 98.6$\%$ & 95.8$\%$ & 91.6$\%$ & 86.2$\%$ & 80.1$\%$ & 4.52 \\
UniVLA~\cite{wang2025unified}            & ABCD$\rightarrow$D & 98.5$\%$ & 96.1$\%$ & 93.1$\%$ & 89.9$\%$ & 85.1$\%$ & 4.63 \\
FLOWER~\cite{reuss2025flower}         & ABCD$\rightarrow$D & \underline{99.2$\%$} & \underline{96.9$\%$} & \textbf{96.9$\%$} & \underline{92.3$\%$} & \underline{88.3$\%$} & \underline{4.67} \\
\rowcolor[HTML]{FFE0CD} 
\texttt{\Ours{}} (Ours) & ABCD$\rightarrow$D & \textbf{99.7$\%$} & \textbf{98.0$\%$} & \underline{96.7$\%$} & \textbf{94.2$\%$} & \textbf{91.8$\%$} & \textbf{4.80} \\
\midrule
RoboFlamingo~\cite{li2023vision}      & ABC$\rightarrow$D & 82.4$\%$ & 61.9$\%$ & 46.6$\%$ & 33.1$\%$ & 23.5$\%$ & 2.48 \\
SuSIE~\cite{black2023zero}            & ABC$\rightarrow$D & 87.0$\%$ & 69.0$\%$ & 49.0$\%$ & 38.0$\%$ & 26.0$\%$ & 2.69 \\
GR-1~\cite{wu2023unleashing}          & ABC$\rightarrow$D & 85.4$\%$ & 71.2$\%$ & 59.6$\%$ & 49.7$\%$ & 40.1$\%$ & 3.06 \\
3DDA~\cite{ke20243d}                  & ABC$\rightarrow$D & 93.8$\%$ & 80.3$\%$ & 66.2$\%$ & 53.3$\%$ & 41.2$\%$ & 3.35 \\
MoDE~\cite{reuss2024efficient}        & ABC$\rightarrow$D & 96.2$\%$ & 88.9$\%$ & 81.1$\%$ & 71.8$\%$ & 63.5$\%$ & 4.01 \\
GR-MG~\cite{li2025gr}                 & ABC$\rightarrow$D & 96.8$\%$ & 89.3$\%$ & 81.5$\%$ & 72.7$\%$ & 64.4$\%$ & 4.04 \\
RoboVLMs~\cite{liu2025towards}        & ABC$\rightarrow$D & 98.0$\%$ & 93.6$\%$ & 85.4$\%$ & 77.8$\%$ & 70.4$\%$ & 4.25 \\
Seer-Large~\cite{tian2024predictive}  & ABC$\rightarrow$D & 96.3$\%$ & 91.6$\%$ & 86.1$\%$ & 80.3$\%$ & 74.0$\%$ & 4.28 \\
VPP~\cite{hu2024video}                & ABC$\rightarrow$D & 95.7$\%$ & 91.2$\%$ & 86.3$\%$ & 81.0$\%$ & 75.0$\%$ & 4.29 \\
UniVLA~\cite{wang2025unified}            & ABC$\rightarrow$D & 98.9$\%$ & 94.8$\%$ & 89.0$\%$ & 82.8$\%$ & 75.1$\%$ & 4.41 \\
FLOWER~\cite{reuss2025flower}         & ABC$\rightarrow$D & \underline{99.4$\%$} & \underline{95.8$\%$} & \underline{90.7$\%$} & \underline{84.9$\%$} & \underline{77.8$\%$} & \underline{4.53} \\
\rowcolor[HTML]{FFE0CD} 
\texttt{\Ours{}} (Ours) & ABC$\rightarrow$D & \textbf{100.0$\%$} & \textbf{98.3$\%$} & \textbf{96.0$\%$} & \textbf{92.6$\%$} & \textbf{88.1$\%$} & \textbf{4.75} \\ \bottomrule
\end{tabular}
\caption{Results on the CALVIN benchmark.}
\label{tab:calvin}
\end{table}

Across all three simulation benchmarks, \Ours{} achieves state-of-the-art (SoTA) performance.
On \textbf{LIBERO} (Tab.~\ref{tab:libero}), using the standard evaluation protocol, we obtain an average success rate of \textbf{98.7\%}, outperforming all the comparing baseline methods.
On \textbf{CALVIN} (Tab.~\ref{tab:calvin}), our method showcases clear advantages in both multi-task long-horizon manipulation (ABCD$\rightarrow$D) and zero-shot environment generalization (ABC$\rightarrow$D).
Measured by the average number of tasks completed in a row of 5, it achieves \textbf{4.80} and \textbf{4.75} in the two settings, respectively, substantially outperforming prior baseline methods.
On \textbf{SimplerEnv} (Tab.~\ref{tab:simplerenv_google_robot} \& \ref{tab:simplerenv_widowx}), \Ours{} achieves average success rates of \textbf{85.5\%} and \textbf{74.7\%} in the Google Robot evaluations under Visual Matching and Variant Aggregation, respectively, as well as \textbf{79.2\%} in the WidowX evaluations, surpassing all comparing baselines.
This consistently strong performance demonstrates robust visual generalization, especially given the substantial visual gap between the real-world training data and the simulated evaluation environments.

\begin{figure}[t]
    \centering
    \includegraphics[width=\linewidth]{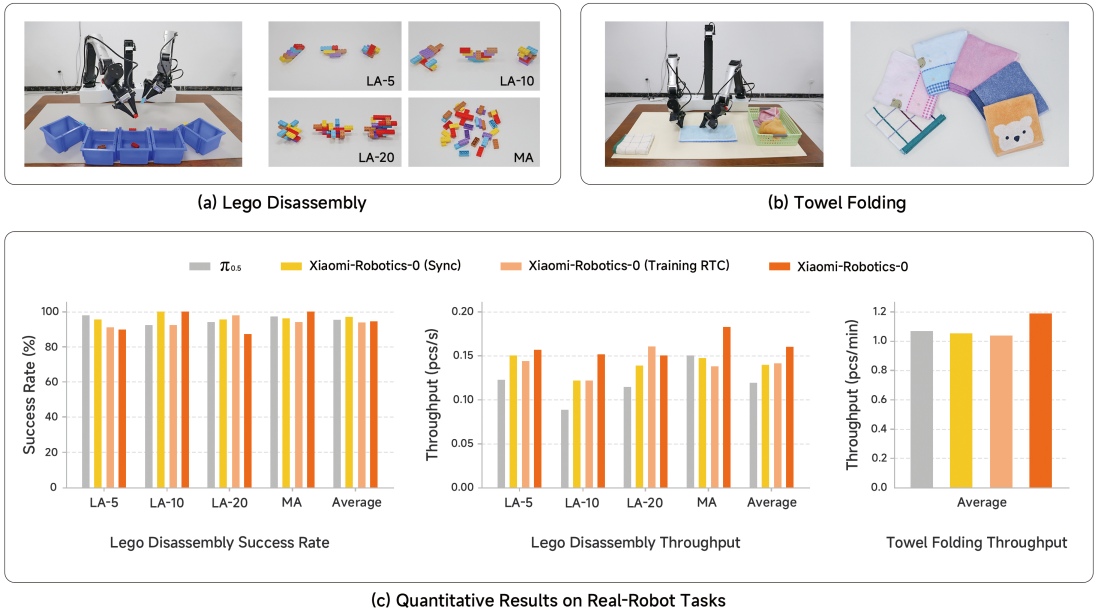}
    \caption{\textbf{Real-Robot Experiments.} (a) We show the setting for Lego Disassembly evaluation. (b) We show the setting for Towel Folding evaluation and all the towels used during evaluation. (c) Quantitative results of different methods on the two tasks.}
    \label{fig:experiment}
\end{figure}

\subsection{Real-Robot Experiments}
\label{experiments:real_robot_experiments}
\subsubsection{Evaluation Details}
To validate real-world performance, we perform experiments with a bimanual robot equipped with two 6-DoF robotic arms (Fig.~\ref{fig:experiment}).
In total, we use three cameras for observation: two wrist-mounted cameras for close-up views and one external camera for a global view.
We evaluate our method on two representative tasks:
\begin{itemize}
    \item \textbf{Lego Disassembly} (Fig.~\ref{fig:experiment}(a)):
    The robot is tasked with 1) disassembling Lego structures into individual bricks, and 2) sorting each brick into the corresponding storage bins according to its color.
    This task requires precise and coordinated bimanual grasping under contact, followed by accurate placement of individual bricks.
    \item \textbf{Towel Folding} (Fig.~\ref{fig:experiment}(b)):
    In this task, the robot needs to 1) pick out a towel from a tray, 2) flatten the towel, 3) fold the towel in half twice, and 4) place the towel to a staging area.
    This task is challenging because towels are deformable and exhibit complex, partially observable dynamics (\textit{e.g.}, wrinkles and occlusions), requiring accurate and coordinated bimanual grasping, and continuous shape control throughout the long-horizon folding sequence.
\end{itemize}

For Lego Disassembly, we evaluate under two settings: large-assembly (LA) and multi-assembly (MA), as illustrated in Fig.~\ref{fig:experiment}(a).
The LA setting evaluates the model's ability to handle increasing complexity within a single structure. 
It includes three sizes, \textit{i.e.}, LA-5, LA-10, and LA-20, which comprises 5, 10, and 20 bricks, respectively.
For each size, we evaluate three different assembly configurations and run three trials per configuration.
The MA setting consists of 34 bricks in total, including both single bricks and groups of two or three bricks assembled together.
We evaluate three trials for this setting.
We report the average success rate, defined as the ratio of correctly sorted bricks to the total number of bricks, as well as the throughput, computed as the number of correctly sorted bricks divided by the total rollout time.
For Towel Folding, we evaluate the policy using six different towels, as shown in Fig.~\ref{fig:experiment}(b). 
We perform two continuous 30-minute rollouts for each method. 
During evaluation, if a single folding attempt exceeds a 2-minute threshold, it is considered a failure. 
We evaluate the performance by reporting the throughput, calculated as the total number of successfully folded towels divided by the rollout duration.

\subsubsection{Implementation Details}
We compare our method with a state-of-the-art baseline and several ablation variants:
\begin{itemize}
    \item $\pi_{0.5}$~\cite{intelligence2025pi_0_5}: A state-of-the-art VLA baseline.
    We follow the official OpenPi\footnote[2]{https://github.com/Physical-Intelligence/openpi} fine-tuning protocol and adapt the released base model to our specific tasks. 
    We use identical training settings as in our experiments.
    \item \Ours{}: Our main method, which incorporates asynchronous execution during post-training to achieve smooth and responsive real-time execution.
    \item \Ours{} (Sync): A synchronous variant of our model for quantifying the performance gains from asynchronous execution.
    \item \Ours{} (Training RTC): The baseline asynchronous variant that leverages training RTC \cite{black2025training} during post-training.
\end{itemize}

We follow the pre-training and post-training procedures described in Sec.~\ref{methods}.
We pre-train the model for 40k steps with a batch size of 32,768. 
We post-train the model with a batch size of 2,048 for 40k steps on Lego Disassembly and 80k steps on Towel Folding.
We use AdamW~\cite{loshchilov2017decoupled} as the optimizer and DeepSpeed ZeRO-2 for training.
We set the action chunk length $T=30$, corresponding to 1 second of actions.

\subsubsection{Results}
Experiment results are summarized in Fig.~\ref{fig:experiment}(c).
In Lego Disassembly, all methods are comparable in terms of average success rates, with the two synchronous methods—$\pi_{0.5}$ and \Ours{} (Sync)—performing slightly better than the other two asynchronous counterparts.
This is because asynchronous methods are less reactive in motion, leading to less precise grasps and high tension between bricks and the gripper fingers, which can cause bricks to eject away from the workspace.
In terms of throughput, \Ours{} (Sync) surpasses $\pi_{0.5}$.
And \Ours{} achieves the highest throughput among all methods, surpassing the training RTC variant which is also deployed asynchronously.
This shows that our proposed post-training techniques are effective in improving execution efficiency in this task that requires high precision.

For Towel Folding, $\pi_{0.5}$, \Ours{} (Sync), and \Ours{} (Training RTC) achieve comparable throughputs of 1 pcs/min.
\Ours{} outperforms these three methods, achieving a throughput of 1.2 pcs/min.
These results demonstrate that our method enables fast execution and robust performance in the challenging deformable object manipulation.
The Training RTC variant often gets stuck when it inadvertantly grasps multiple layers of the towel during the flinging motion, preventing the motion from flattening the towel.
Rather than re-grasping to correct this, the policy falls into a repetitive loop, repeatedly executing the flinging motion.
This observation suggests that the action-prefixing mechanism introduces a shortcut in policy learning, allowing later-timestep action predictions to simply copy the prefixed actions rather than attend to signals from other modalities.
In contrast, \Ours{} is able to effectively avoid such repetitive failures.

\subsection{Preservation of Vision-Language Capabilities}
\label{experiments:preservation}
During pre-training, we jointly train \Ours{} on both vision-language data and robot trajectory data (Sec.~\ref{methods:model_training}).
This enables the model to avoid catastrophic forgetting of the vision-language capabilities of the underlying pre-trained VLM and enhance visual perception on robot-centric data.
To validate this, we evaluate the VLM of \Ours{} after pre-training on a comprehensive suite of vision-language (VL) benchmarks~\cite{liu2024mmbench, ying-etal-2025-seedbench, Li-hallucination-2023, fu2025mme, team2025gemini, lu2022learn, yue2023mmmu, kembhavi2016diagram, Orogat2021, singh2019towards}, covering tasks from general QA to hallucination detection.
We further report results on ERQA~\cite{team2025gemini}, a benchmark designed for evaluating embodied reasoning capabilities.
We compare with two state-of-the-art vision-language-action (VLA) models, $\pi_{0.5}$~\cite{intelligence2025pi_0_5} and MolmoAct~\cite{lee2025molmoact}, which also incorporate vision-language data during training.
We also include comparison with VLA models that \textit{do not} incorporate VL data during training to understand how the VL capabilities degrade in the absence of the corresponding data.
Specifically, we compare with $\pi_0$~\cite{black2024pi_0} and a variant of our method which removes VL data during pre-training (denoted as \Ours{} (w/o VL data)).

\begin{table}[ht]
\centering
\resizebox{\textwidth}{!}{
\begin{tabular}{lcccccccccc}
\toprule
\textbf{Model} & \textbf{ERQA} & \textbf{SEED} & \textbf{POPE} & \textbf{AI2D} & \textbf{MMBench} & \textbf{MME} & \textbf{MMMU} & \textbf{TextVQA} & \textbf{SciQA} & \textbf{ChartQA} \\
\midrule
$\pi_{0}$~\cite{black2024pi_0} & 0.0 & 0.0 & 0.0 & 0.0 & 0.0 & 0.1 & 0.1 & 1.4 & 0.0 & 0.0 \\
$\pi_{0.5}$~\cite{intelligence2025pi_0_5} & 0.0 & 21.5 & 0.0 & 14.4 & 22.1 & 0.0 & 19.9 & 0.0 & 28.0 & 0.5 \\
MolmoAct~\cite{lee2025molmoact} & \underline{33.5} & \underline{72.7} & \underline{86.6} & \underline{72.0} & \underline{80.1} & \underline{69.5} & \underline{38.0} & \underline{67.3} & \textbf{91.1} & \underline{57.1} \\
\Ours{} (w/o VL data) & 0.0 & 0.0 & 0.0 & 0.0 & 0.0 & 0.0 & 0.0 & 0.0 & 0.0 & 0.0 \\
\rowcolor[HTML]{FFE0CD}
\Ours{} (Ours) & \textbf{40.8} & \textbf{78.6} & \textbf{88.5} & \textbf{78.7} & \textbf{84.4} & \textbf{81.8} & \textbf{46.2} & \textbf{72.0} & \underline{79.4} & \textbf{59.2} \\
\midrule
Qwen3-VL-4B-Instruct~\cite{bai2025qwen3} & 40.0 & 78.8 & 89.7 & 81.6 & 88.7 & 87.1 & 51.7 & 78.0 & 92.7 & 76.8 \\
\bottomrule
\end{tabular}
}
\caption{
Quantitative results on general vision-language and embodied reasoning benchmarks.
See App.~\ref{app:benchmark_details} for detailed definitions and evaluation metrics for each benchmark.
}
\label{tab:vlm_benchmark}
\end{table}

Results are shown in Tab.~\ref{tab:vlm_benchmark}.
\Ours{} outperforms all the comparing VLA baselines in all but one benchmarks.
It is able to effectively preserve the vision-language capabilities of the underlying pre-trained VLM, trailing slightly behind it on most general VL benchmarks.
It showcases strong performance in the challenging object hallucination evaluations (the POPE series) and OCR-related tasks (\textit{e.g.}, AI2D).
Surprisingly, \Ours{} slightly surpasses Qwen3-VL-4B-Instruct on the ERQA benchmark (40.8 v.s. 40.0).
We hypothesize that this gain stems from incorporating vision-language data derived from robot trajectory data into the training mixture, which strengthens visual perception on robot-centric inputs.
$\pi_0$ achieves near-zero performance on most VL tasks, while \Ours{} (w/o VL data) attains zero performance across all tasks. 
These results indicate that without explicit vision-language supervision, training on robot trajectories alone fails to retain the general-purpose VL knowledge, leading to severe catastrophic forgetting.
We provide additional qualitative results of \Ours{} on different evaluation benchmarks in App.~\ref{app:vl_competence_demos}.

\section{Related Work}
\label{related_work}
Recently, vision-language-action (VLA) models have emerged as a new paradigm for robot policy learning~\cite{brohan2022rt, zitkovich2023rt2, black2024pi_0, intelligence2025pi_0_5, pertsch2025fast, liu2025towards, gr00tn1_2025, cheang2025gr, kim2024openvla, kim2025fine, bu2025univla}.
By leveraging large-scale robot data collected across diverse embodiments, tasks, and environments~\cite{Neill2023open_x_embodiment, bu2025agibot, khazatsky2024droid, walke2023bridgedata}, VLA models can effectively solve a broad range of tasks and showcase strong generalization capabilities on handling various kinds of out-of-distribution settings, including novel environments, instructions, and objects~\cite{zitkovich2023rt2, intelligence2025pi_0_5, cheang2025gr}.
Typically, VLAs are built upon pre-trained vision-language models (VLMs) that have been trained to capture broad visual semantic knowledge.
A straightforward approach is to convert actions to discretized tokens and train the VLM to generate actions via a next-token-prediction objective~\cite{zitkovich2023rt2, kim2024openvla, pertsch2025fast}.
However, action tokenization can introduce quantization error and reduce control precision.
Another effective method of modeling the complex trajectory distribution is to leverage the expressive power of flow matching~\cite{black2024pi_0, intelligence2025pi_0_5, gr00tn1_2025, cheang2025gr, qu2025eo} or diffusion~\cite{liu2024rdt, liu2025hybridvla, liu2026rdt2, chi2025diffusion, wen2025dexvla}.
To effectively transfer knowledge from the pre-trained VLMs to VLA models, it is crucial to preserve the VLMs’ capabilities throughout training the VLA models.
One simple method is to jointly train on both vision-language (VL) data and robot trajectories~\cite{lee2025molmoact, intelligence2025pi_0_5, zitkovich2023rt2}.
Recent work \cite{driess2025knowledge} further proposes detaching the flow-matching objective from the VLM backbone, thereby preventing gradient backpropagation into it.
We build \Ours{} by integrating a pre-trained VLM~\cite{bai2025qwen3} with a diffusion transformer~\cite{peebles2023scalable} that generates actions via flow matching.
During pre-training, we jointly train on both robot trajectories and VL data in the first stage and freeze the VLM while training the DiT in the second stage to avoid catastrophic forgetting of the vision-language knowledge.

Given the large number of parameters—often up to billions—the inference latency of large VLA models is non-negligible during real-robot rollouts~\cite{black2025real}.
A simple execution strategy is to roll out policies synchronously~\cite{black2024pi_0}, where the robot remains idle until the next inference completes, causing pauses and discontinuous actions.
Another method is to accelerate inference to achieve real-time performance~\cite{ma2025running}.
Recent work explores asynchronous execution, where the robot continues executing while the model performs inference~\cite{black2024pi_0, intelligence2025pi_0_5, tang2025vlash}.
A line of work proposes to prefix previously generated and committed actions in the prediction of the next action chunk.
Real-Time Chunking (RTC)~\cite{black2025real} leverages a training-free inpainting algorithm that ``freezes'' the prefixed action and ``inpaints'' the rest in a way that is consistent with the frozen prefix.
Training RTC~\cite{black2025training} incorporates the prefixed actions during training.
However, conditioning on prefixed actions during training allows later-timestep predictions to leverage the shortcut of exploiting temporal correlations between consecutive actions, resulting in less reactive behavior.
In this work, we propose several practical techniques to address this issue, achieving high throughput on challenging tasks that require precise and dexterous manipulation.

\section{Conclusions}
\label{conclusions}
We introduce \texttt{\Ours}, a powerful vision-language-action (VLA) model designed for both high performance and smooth real-time execution.
\texttt{\Ours} is pre-trained on large-scale robot trajectories and vision-language data, enabling strong action generation capabilities while preventing catastrophic forgetting of the visual-semantic knowledge in the underlying pre-trained VLM.
In post-training, we develop several practical techniques to train the VLA model for asynchronous execution, allowing continuous and reactive real-time execution on real robots.
We evaluate \texttt{\Ours} on extensive simulation benchmarks and two real-robot tasks requiring precise and dexterous bimanual manipulation.
Results showcase that the proposed method delivers state-of-the-art performance on all simulation benchmarks.
In addition, \texttt{\Ours} runs fast and smoothly on real robots with a consumer-grade GPU, achieving high success rates and strong throughput in both real-world tasks.
In the future, we plan to explore training the model on larger and more diverse robot datasets and continue to improve its robustness and generalization capabilities in real-world tasks.

\clearpage

\section*{Contributions}
\label{contributions}
Authors are listed in alphabetical order. 

Rui Cai, Jun Guo, Xinze He, Piaopiao Jin, Jie Li, Bingxuan Lin, Futeng Liu, Wei Liu, Fei Ma, Kun Ma, Feng Qiu, Heng Qu, Yifei Su, Qiao Sun, Dong Wang, Donghao Wang, Yunhong Wang, Rujie Wu, Diyun Xiang, Yu Yang, Hangjun Ye, Yuan Zhang, Quanyun Zhou

\clearpage

\bibliographystyle{plainnat}
\bibliography{main}

\begin{thebibliography}{76}
\providecommand{\natexlab}[1]{#1}
\providecommand{\url}[1]{\texttt{#1}}
\expandafter\ifx\csname urlstyle\endcsname\relax
  \providecommand{\doi}[1]{doi: #1}\else
  \providecommand{\doi}{doi: \begingroup \urlstyle{rm}\Url}\fi

\bibitem[Bai et~al.(2025)Bai, Cai, Chen, Chen, Chen, Cheng, Deng, Ding, Gao, Ge, et~al.]{bai2025qwen3}
Shuai Bai, Yuxuan Cai, Ruizhe Chen, Keqin Chen, Xionghui Chen, Zesen Cheng, Lianghao Deng, Wei Ding, Chang Gao, Chunjiang Ge, et~al.
\newblock {Qwen3-VL} technical report.
\newblock \emph{arXiv preprint arXiv:2511.21631}, 2025.

\bibitem[Black et~al.(2023)Black, Nakamoto, Atreya, Walke, Finn, Kumar, and Levine]{black2023zero}
Kevin Black, Mitsuhiko Nakamoto, Pranav Atreya, Homer Walke, Chelsea Finn, Aviral Kumar, and Sergey Levine.
\newblock Zero-shot robotic manipulation with pretrained image-editing diffusion models.
\newblock \emph{arXiv preprint arXiv:2310.10639}, 2023.

\bibitem[Black et~al.(2024)Black, Brown, Driess, Esmail, Equi, Finn, Fusai, Groom, Hausman, Ichter, et~al.]{black2024pi_0}
Kevin Black, Noah Brown, Danny Driess, Adnan Esmail, Michael Equi, Chelsea Finn, Niccolo Fusai, Lachy Groom, Karol Hausman, Brian Ichter, et~al.
\newblock {$\pi_0$: A Vision-Language-Action Flow Model for General Robot Control}.
\newblock \emph{arXiv preprint arXiv:2410.24164}, 2024.

\bibitem[Black et~al.(2025{\natexlab{a}})Black, Galliker, and Levine]{black2025real}
Kevin Black, Manuel~Y Galliker, and Sergey Levine.
\newblock Real-time execution of action chunking flow policies.
\newblock \emph{arXiv preprint arXiv:2506.07339}, 2025{\natexlab{a}}.

\bibitem[Black et~al.(2025{\natexlab{b}})Black, Ren, Equi, and Levine]{black2025training}
Kevin Black, Allen~Z Ren, Michael Equi, and Sergey Levine.
\newblock Training-time action conditioning for efficient real-time chunking.
\newblock \emph{arXiv preprint arXiv:2512.05964}, 2025{\natexlab{b}}.

\bibitem[Brohan et~al.(2022)Brohan, Brown, Carbajal, Chebotar, Dabis, Finn, Gopalakrishnan, Hausman, Herzog, Hsu, et~al.]{brohan2022rt}
Anthony Brohan, Noah Brown, Justice Carbajal, Yevgen Chebotar, Joseph Dabis, Chelsea Finn, Keerthana Gopalakrishnan, Karol Hausman, Alex Herzog, Jasmine Hsu, et~al.
\newblock {RT-1}: Robotics transformer for real-world control at scale.
\newblock \emph{arXiv preprint arXiv:2212.06817}, 2022.

\bibitem[Bu et~al.(2025{\natexlab{a}})Bu, Cai, Chen, Cui, Ding, Feng, Gao, He, Hu, Huang, et~al.]{bu2025agibot}
Qingwen Bu, Jisong Cai, Li~Chen, Xiuqi Cui, Yan Ding, Siyuan Feng, Shenyuan Gao, Xindong He, Xuan Hu, Xu~Huang, et~al.
\newblock {AgiBot} world colosseo: A large-scale manipulation platform for scalable and intelligent embodied systems.
\newblock \emph{arXiv preprint arXiv:2503.06669}, 2025{\natexlab{a}}.

\bibitem[Bu et~al.(2025{\natexlab{b}})Bu, Yang, Cai, Gao, Ren, Yao, Luo, and Li]{bu2025univla}
Qingwen Bu, Yanting Yang, Jisong Cai, Shenyuan Gao, Guanghui Ren, Maoqing Yao, Ping Luo, and Hongyang Li.
\newblock {UniVLA}: Learning to act anywhere with task-centric latent actions.
\newblock \emph{arXiv preprint arXiv:2505.06111}, 2025{\natexlab{b}}.

\bibitem[Changpinyo et~al.(2021)Changpinyo, Sharma, Ding, and Soricut]{changpinyo2021cc12m}
Soravit Changpinyo, Piyush Sharma, Nan Ding, and Radu Soricut.
\newblock {Conceptual 12M}: Pushing web-scale image-text pre-training to recognize long-tail visual concepts.
\newblock In \emph{CVPR}, 2021.

\bibitem[Cheang et~al.(2025)Cheang, Chen, Cui, Hu, Huang, Kong, Li, Li, Liu, Ma, et~al.]{cheang2025gr}
Chilam Cheang, Sijin Chen, Zhongren Cui, Yingdong Hu, Liqun Huang, Tao Kong, Hang Li, Yifeng Li, Yuxiao Liu, Xiao Ma, et~al.
\newblock {GR-3} technical report.
\newblock \emph{arXiv preprint arXiv:2507.15493}, 2025.

\bibitem[Chi et~al.(2025)Chi, Xu, Feng, Cousineau, Du, Burchfiel, Tedrake, and Song]{chi2025diffusion}
Cheng Chi, Zhenjia Xu, Siyuan Feng, Eric Cousineau, Yilun Du, Benjamin Burchfiel, Russ Tedrake, and Shuran Song.
\newblock Diffusion policy: Visuomotor policy learning via action diffusion.
\newblock \emph{The International Journal of Robotics Research}, 44\penalty0 (10-11):\penalty0 1684--1704, 2025.

\bibitem[Collaboration et~al.(2023)Collaboration, O'Neill, Rehman, Gupta, Maddukuri, Gupta, Padalkar, Lee, Pooley, Gupta, Mandlekar, Jain, Tung, Bewley, Herzog, Irpan, Khazatsky, Rai, Gupta, Wang, Kolobov, Singh, Garg, Kembhavi, Xie, Brohan, Raffin, Sharma, Yavary, Jain, Balakrishna, Wahid, Burgess-Limerick, Kim, Schölkopf, Wulfe, Ichter, Lu, Xu, Le, Finn, Wang, Xu, Chi, Huang, Chan, Agia, Pan, Fu, Devin, Xu, Morton, Driess, Chen, Pathak, Shah, Büchler, Jayaraman, Kalashnikov, Sadigh, Johns, Foster, Liu, Ceola, Xia, Zhao, Frujeri, Stulp, Zhou, Sukhatme, Salhotra, Yan, Feng, Schiavi, Berseth, Kahn, Yang, Wang, Su, Fang, Shi, Bao, Amor, Christensen, Furuta, Bharadhwaj, Walke, Fang, Ha, Mordatch, Radosavovic, Leal, Liang, Abou-Chakra, Kim, Drake, Peters, Schneider, Hsu, Vakil, Bohg, Bingham, Wu, Gao, Hu, Wu, Wu, Sun, Luo, Gu, Tan, Oh, Wu, Lu, Yang, Malik, Silvério, Hejna, Booher, Tompson, Yang, Salvador, Lim, Han, Wang, Rao, Pertsch, Hausman, Go, Gopalakrishnan, Goldberg, Byrne, Oslund, Kawaharazuka, Black,
  Lin, Zhang, Ehsani, Lekkala, Ellis, Rana, Srinivasan, Fang, Singh, Zeng, Hatch, Hsu, Itti, Chen, Pinto, Fei-Fei, Tan, Fan, Ott, Lee, Weihs, Chen, Lepert, Memmel, Tomizuka, Itkina, Castro, Spero, Du, Ahn, Yip, Zhang, Ding, Heo, Srirama, Sharma, Kim, Irshad, Kanazawa, Hansen, Heess, Joshi, Suenderhauf, Liu, Palo, Shafiullah, Mees, Kroemer, Bastani, Sanketi, Miller, Yin, Wohlhart, Xu, Fagan, Mitrano, Sermanet, Abbeel, Sundaresan, Chen, Vuong, Rafailov, Tian, Doshi, Mart{'i}n-Mart{'i}n, Baijal, Scalise, Hendrix, Lin, Qian, Zhang, Mendonca, Shah, Hoque, Julian, Bustamante, Kirmani, Levine, Lin, Moore, Bahl, Dass, Sonawani, Tulsiani, Song, Xu, Haldar, Karamcheti, Adebola, Guist, Nasiriany, Schaal, Welker, Tian, Ramamoorthy, Dasari, Belkhale, Park, Nair, Mirchandani, Osa, Gupta, Harada, Matsushima, Xiao, Kollar, Yu, Ding, Davchev, Zhao, Armstrong, Darrell, Chung, Jain, Kumar, Vanhoucke, Guizilini, Zhan, Zhou, Burgard, Chen, Chen, Wang, Zhu, Geng, Liu, Liangwei, Li, Pang, Lu, Ma, Kim, Chebotar, Zhou, Zhu, Wu, Xu,
  Wang, Bisk, Dou, Cho, Lee, Cui, Cao, Wu, Tang, Zhu, Zhang, Jiang, Li, Li, Iwasawa, Matsuo, Ma, Xu, Cui, Zhang, Fu, and Lin]{Neill2023open_x_embodiment}
Open X-Embodiment Collaboration, Abby O'Neill, Abdul Rehman, Abhinav Gupta, Abhiram Maddukuri, Abhishek Gupta, Abhishek Padalkar, Abraham Lee, Acorn Pooley, Agrim Gupta, Ajay Mandlekar, Ajinkya Jain, Albert Tung, Alex Bewley, Alex Herzog, Alex Irpan, Alexander Khazatsky, Anant Rai, Anchit Gupta, Andrew Wang, Andrey Kolobov, Anikait Singh, Animesh Garg, Aniruddha Kembhavi, Annie Xie, Anthony Brohan, Antonin Raffin, Archit Sharma, Arefeh Yavary, Arhan Jain, Ashwin Balakrishna, Ayzaan Wahid, Ben Burgess-Limerick, Beomjoon Kim, Bernhard Schölkopf, Blake Wulfe, Brian Ichter, Cewu Lu, Charles Xu, Charlotte Le, Chelsea Finn, Chen Wang, Chenfeng Xu, Cheng Chi, Chenguang Huang, Christine Chan, Christopher Agia, Chuer Pan, Chuyuan Fu, Coline Devin, Danfei Xu, Daniel Morton, Danny Driess, Daphne Chen, Deepak Pathak, Dhruv Shah, Dieter Büchler, Dinesh Jayaraman, Dmitry Kalashnikov, Dorsa Sadigh, Edward Johns, Ethan Foster, Fangchen Liu, Federico Ceola, Fei Xia, Feiyu Zhao, Felipe~Vieira Frujeri, Freek Stulp, Gaoyue
  Zhou, Gaurav~S. Sukhatme, Gautam Salhotra, Ge~Yan, Gilbert Feng, Giulio Schiavi, Glen Berseth, Gregory Kahn, Guangwen Yang, Guanzhi Wang, Hao Su, Hao-Shu Fang, Haochen Shi, Henghui Bao, Heni~Ben Amor, Henrik~I Christensen, Hiroki Furuta, Homanga Bharadhwaj, Homer Walke, Hongjie Fang, Huy Ha, Igor Mordatch, Ilija Radosavovic, Isabel Leal, Jacky Liang, Jad Abou-Chakra, Jaehyung Kim, Jaimyn Drake, Jan Peters, Jan Schneider, Jasmine Hsu, Jay Vakil, Jeannette Bohg, Jeffrey Bingham, Jeffrey Wu, Jensen Gao, Jiaheng Hu, Jiajun Wu, Jialin Wu, Jiankai Sun, Jianlan Luo, Jiayuan Gu, Jie Tan, Jihoon Oh, Jimmy Wu, Jingpei Lu, Jingyun Yang, Jitendra Malik, João Silvério, Joey Hejna, Jonathan Booher, Jonathan Tompson, Jonathan Yang, Jordi Salvador, Joseph~J. Lim, Junhyek Han, Kaiyuan Wang, Kanishka Rao, Karl Pertsch, Karol Hausman, Keegan Go, Keerthana Gopalakrishnan, Ken Goldberg, Kendra Byrne, Kenneth Oslund, Kento Kawaharazuka, Kevin Black, Kevin Lin, Kevin Zhang, Kiana Ehsani, Kiran Lekkala, Kirsty Ellis, Krishan
  Rana, Krishnan Srinivasan, Kuan Fang, Kunal~Pratap Singh, Kuo-Hao Zeng, Kyle Hatch, Kyle Hsu, Laurent Itti, Lawrence~Yunliang Chen, Lerrel Pinto, Li~Fei-Fei, Liam Tan, Linxi~"Jim" Fan, Lionel Ott, Lisa Lee, Luca Weihs, Magnum Chen, Marion Lepert, Marius Memmel, Masayoshi Tomizuka, Masha Itkina, Mateo~Guaman Castro, Max Spero, Maximilian Du, Michael Ahn, Michael~C. Yip, Mingtong Zhang, Mingyu Ding, Minho Heo, Mohan~Kumar Srirama, Mohit Sharma, Moo~Jin Kim, Muhammad~Zubair Irshad, Naoaki Kanazawa, Nicklas Hansen, Nicolas Heess, Nikhil~J Joshi, Niko Suenderhauf, Ning Liu, Norman~Di Palo, Nur Muhammad~Mahi Shafiullah, Oier Mees, Oliver Kroemer, Osbert Bastani, Pannag~R Sanketi, Patrick~"Tree" Miller, Patrick Yin, Paul Wohlhart, Peng Xu, Peter~David Fagan, Peter Mitrano, Pierre Sermanet, Pieter Abbeel, Priya Sundaresan, Qiuyu Chen, Quan Vuong, Rafael Rafailov, Ran Tian, Ria Doshi, Roberto Mart{'i}n-Mart{'i}n, Rohan Baijal, Rosario Scalise, Rose Hendrix, Roy Lin, Runjia Qian, Ruohan Zhang, Russell Mendonca, Rutav
  Shah, Ryan Hoque, Ryan Julian, Samuel Bustamante, Sean Kirmani, Sergey Levine, Shan Lin, Sherry Moore, Shikhar Bahl, Shivin Dass, Shubham Sonawani, Shubham Tulsiani, Shuran Song, Sichun Xu, Siddhant Haldar, Siddharth Karamcheti, Simeon Adebola, Simon Guist, Soroush Nasiriany, Stefan Schaal, Stefan Welker, Stephen Tian, Subramanian Ramamoorthy, Sudeep Dasari, Suneel Belkhale, Sungjae Park, Suraj Nair, Suvir Mirchandani, Takayuki Osa, Tanmay Gupta, Tatsuya Harada, Tatsuya Matsushima, Ted Xiao, Thomas Kollar, Tianhe Yu, Tianli Ding, Todor Davchev, Tony~Z. Zhao, Travis Armstrong, Trevor Darrell, Trinity Chung, Vidhi Jain, Vikash Kumar, Vincent Vanhoucke, Vitor Guizilini, Wei Zhan, Wenxuan Zhou, Wolfram Burgard, Xi~Chen, Xiangyu Chen, Xiaolong Wang, Xinghao Zhu, Xinyang Geng, Xiyuan Liu, Xu~Liangwei, Xuanlin Li, Yansong Pang, Yao Lu, Yecheng~Jason Ma, Yejin Kim, Yevgen Chebotar, Yifan Zhou, Yifeng Zhu, Yilin Wu, Ying Xu, Yixuan Wang, Yonatan Bisk, Yongqiang Dou, Yoonyoung Cho, Youngwoon Lee, Yuchen Cui, Yue Cao,
  Yueh-Hua Wu, Yujin Tang, Yuke Zhu, Yunchu Zhang, Yunfan Jiang, Yunshuang Li, Yunzhu Li, Yusuke Iwasawa, Yutaka Matsuo, Zehan Ma, Zhuo Xu, Zichen~Jeff Cui, Zichen Zhang, Zipeng Fu, and Zipeng Lin.
\newblock Open {X-E}mbodiment: Robotic learning datasets and {RT-X} models.
\newblock \url{https://arxiv.org/abs/2310.08864}, 2023.

\bibitem[Driess et~al.(2025)Driess, Springenberg, Ichter, Yu, Li-Bell, Pertsch, Ren, Walke, Vuong, Shi, et~al.]{driess2025knowledge}
Danny Driess, Jost~Tobias Springenberg, Brian Ichter, Lili Yu, Adrian Li-Bell, Karl Pertsch, Allen~Z Ren, Homer Walke, Quan Vuong, Lucy~Xiaoyang Shi, et~al.
\newblock Knowledge insulating vision-language-action models: Train fast, run fast, generalize better.
\newblock \emph{arXiv preprint arXiv:2505.23705}, 2025.

\bibitem[Fu et~al.(2025{\natexlab{a}})Fu, Chen, Shen, Qin, Zhang, Lin, Yang, Zheng, Li, Sun, et~al.]{fu2025mme}
Chaoyou Fu, Peixian Chen, Yunhang Shen, Yulei Qin, Mengdan Zhang, Xu~Lin, Jinrui Yang, Xiawu Zheng, Ke~Li, Xing Sun, et~al.
\newblock {MME}: A comprehensive evaluation benchmark for multimodal large language models.
\newblock In \emph{The Thirty-ninth Annual Conference on Neural Information Processing Systems Datasets and Benchmarks Track}, 2025{\natexlab{a}}.

\bibitem[Fu et~al.(2025{\natexlab{b}})Fu, Yang, Mo, Yan, Wei, Meng, Xie, and Zheng]{fu2025llmdet}
Shenghao Fu, Qize Yang, Qijie Mo, Junkai Yan, Xihan Wei, Jingke Meng, Xiaohua Xie, and Wei-Shi Zheng.
\newblock {LLMDet}: Learning strong open-vocabulary object detectors under the supervision of large language models.
\newblock \emph{arXiv preprint arXiv:2501.18954}, 2025{\natexlab{b}}.

\bibitem[Han et~al.(2024)Han, Wang, Peng, Xiong, Chen, Ji, and Wang]{han2024lm}
Chi Han, Qifan Wang, Hao Peng, Wenhan Xiong, Yu~Chen, Heng Ji, and Sinong Wang.
\newblock {LM-Infinite}: Zero-shot extreme length generalization for large language models.
\newblock In \emph{Proceedings of the 2024 Conference of the North American Chapter of the Association for Computational Linguistics: Human Language Technologies (Volume 1: Long Papers)}, pages 3991--4008, 2024.

\bibitem[Hu et~al.(2024)Hu, Guo, Wang, Chen, Wang, Zhang, Sreenath, Lu, and Chen]{hu2024video}
Yucheng Hu, Yanjiang Guo, Pengchao Wang, Xiaoyu Chen, Yen-Jen Wang, Jianke Zhang, Koushil Sreenath, Chaochao Lu, and Jianyu Chen.
\newblock {Video Prediction Policy}: A generalist robot policy with predictive visual representations.
\newblock \emph{arXiv preprint arXiv:2412.14803}, 2024.

\bibitem[Huang et~al.(2025)Huang, Wu, Chen, Wang, and Yang]{huang2025thinkact}
Chi-Pin Huang, Yueh-Hua Wu, Min-Hung Chen, Yu-Chiang~Frank Wang, and Fu-En Yang.
\newblock {ThinkAct}: Vision-language-action reasoning via reinforced visual latent planning.
\newblock \emph{arXiv preprint arXiv:2507.16815}, 2025.

\bibitem[Intelligence et~al.(2025)Intelligence, Black, Brown, Darpinian, Dhabalia, Driess, Esmail, Equi, Finn, Fusai, et~al.]{intelligence2025pi_0_5}
Physical Intelligence, Kevin Black, Noah Brown, James Darpinian, Karan Dhabalia, Danny Driess, Adnan Esmail, Michael Equi, Chelsea Finn, Niccolo Fusai, et~al.
\newblock {$\pi_{0.5}$: A Vision-Language-Action Model with Open-World Generalization}.
\newblock \emph{arXiv preprint arXiv:2504.16054}, 2025.

\bibitem[Jiang et~al.(2024)Jiang, Li, Zhang, Wu, Luo, Ahn, Han, Abdi, Li, Lin, et~al.]{jiang2024minference}
Huiqiang Jiang, Yucheng Li, Chengruidong Zhang, Qianhui Wu, Xufang Luo, Surin Ahn, Zhenhua Han, Amir~H Abdi, Dongsheng Li, Chin-Yew Lin, et~al.
\newblock {MInference} 1.0: Accelerating pre-filling for long-context llms via dynamic sparse attention.
\newblock \emph{Advances in Neural Information Processing Systems}, 37:\penalty0 52481--52515, 2024.

\bibitem[Ke et~al.(2024)Ke, Gkanatsios, and Fragkiadaki]{ke20243d}
Tsung-Wei Ke, Nikolaos Gkanatsios, and Katerina Fragkiadaki.
\newblock {3D Diffuser Actor}: Policy diffusion with 3d scene representations.
\newblock \emph{arXiv preprint arXiv:2402.10885}, 2024.

\bibitem[Kembhavi et~al.(2016)Kembhavi, Salvato, Kolve, Seo, Hajishirzi, and Farhadi]{kembhavi2016diagram}
Aniruddha Kembhavi, Mike Salvato, Eric Kolve, Minjoon Seo, Hannaneh Hajishirzi, and Ali Farhadi.
\newblock A diagram is worth a dozen images.
\newblock In \emph{European conference on computer vision}, pages 235--251. Springer, 2016.

\bibitem[Khazatsky et~al.(2024)Khazatsky, Pertsch, Nair, Balakrishna, Dasari, Karamcheti, Nasiriany, Srirama, Chen, Ellis, et~al.]{khazatsky2024droid}
Alexander Khazatsky, Karl Pertsch, Suraj Nair, Ashwin Balakrishna, Sudeep Dasari, Siddharth Karamcheti, Soroush Nasiriany, Mohan~Kumar Srirama, Lawrence~Yunliang Chen, Kirsty Ellis, et~al.
\newblock {DROID}: A large-scale in-the-wild robot manipulation dataset.
\newblock \emph{arXiv preprint arXiv:2403.12945}, 2024.

\bibitem[Kim et~al.(2024)Kim, Pertsch, Karamcheti, Xiao, Balakrishna, Nair, Rafailov, Foster, Lam, Sanketi, et~al.]{kim2024openvla}
Moo~Jin Kim, Karl Pertsch, Siddharth Karamcheti, Ted Xiao, Ashwin Balakrishna, Suraj Nair, Rafael Rafailov, Ethan Foster, Grace Lam, Pannag Sanketi, et~al.
\newblock {OpenVLA}: An open-source vision-language-action model.
\newblock \emph{arXiv preprint arXiv:2406.09246}, 2024.

\bibitem[Kim et~al.(2025)Kim, Finn, and Liang]{kim2025fine}
Moo~Jin Kim, Chelsea Finn, and Percy Liang.
\newblock Fine-tuning vision-language-action models: Optimizing speed and success.
\newblock \emph{arXiv preprint arXiv:2502.19645}, 2025.

\bibitem[Lee et~al.(2025)Lee, Duan, Fang, Deng, Liu, Li, Fang, Zhang, Wang, Lee, et~al.]{lee2025molmoact}
Jason Lee, Jiafei Duan, Haoquan Fang, Yuquan Deng, Shuo Liu, Boyang Li, Bohan Fang, Jieyu Zhang, Yi~Ru Wang, Sangho Lee, et~al.
\newblock {MolmoAct}: Action reasoning models that can reason in space.
\newblock \emph{arXiv preprint arXiv:2508.07917}, 2025.

\bibitem[Li et~al.(2025)Li, Wu, Huang, Cheang, Wang, and Kong]{li2025gr}
Peiyan Li, Hongtao Wu, Yan Huang, Chilam Cheang, Liang Wang, and Tao Kong.
\newblock {GR-MG}: Leveraging partially annotated data via multi-modal goal-conditioned policy.
\newblock \emph{IEEE Robotics and Automation Letters}, 2025.

\bibitem[Li et~al.(2023{\natexlab{a}})Li, Liu, Zhang, Yu, Xu, Wu, Cheang, Jing, Zhang, Liu, et~al.]{li2023vision}
Xinghang Li, Minghuan Liu, Hanbo Zhang, Cunjun Yu, Jie Xu, Hongtao Wu, Chilam Cheang, Ya~Jing, Weinan Zhang, Huaping Liu, et~al.
\newblock Vision-language foundation models as effective robot imitators.
\newblock \emph{arXiv preprint arXiv:2311.01378}, 2023{\natexlab{a}}.

\bibitem[Li et~al.(2024{\natexlab{a}})Li, Li, Liu, Wang, Liu, Kang, Ma, Kong, Zhang, and Liu]{liu2025towards}
Xinghang Li, Peiyan Li, Minghuan Liu, Dong Wang, Jirong Liu, Bingyi Kang, Xiao Ma, Tao Kong, Hanbo Zhang, and Huaping Liu.
\newblock Towards generalist robot policies: What matters in building vision-language-action models.
\newblock \emph{arXiv preprint arXiv:2412.14058}, 2024{\natexlab{a}}.

\bibitem[Li et~al.(2024{\natexlab{b}})Li, Hsu, Gu, Pertsch, Mees, Walke, Fu, Lunawat, Sieh, Kirmani, et~al.]{li2024evaluating}
Xuanlin Li, Kyle Hsu, Jiayuan Gu, Karl Pertsch, Oier Mees, Homer~Rich Walke, Chuyuan Fu, Ishikaa Lunawat, Isabel Sieh, Sean Kirmani, et~al.
\newblock Evaluating real-world robot manipulation policies in simulation.
\newblock \emph{arXiv preprint arXiv:2405.05941}, 2024{\natexlab{b}}.

\bibitem[Li et~al.(2023{\natexlab{b}})Li, Du, Zhou, Wang, Zhao, and Wen]{Li-hallucination-2023}
Yifan Li, Yifan Du, Kun Zhou, Jinpeng Wang, Wayne~Xin Zhao, and Ji-Rong Wen.
\newblock Evaluating object hallucination in large vision-language models.
\newblock In \emph{The 2023 Conference on Empirical Methods in Natural Language Processing}, 2023{\natexlab{b}}.
\newblock URL \url{https://openreview.net/forum?id=xozJw0kZXF}.

\bibitem[Liang et~al.(2024)Liang, Yu, Luo, Iyer, Dong, Zhou, Ghosh, Lewis, Yih, Zettlemoyer, et~al.]{liang2024mixture}
Weixin Liang, Lili Yu, Liang Luo, Srinivasan Iyer, Ning Dong, Chunting Zhou, Gargi Ghosh, Mike Lewis, Wen-tau Yih, Luke Zettlemoyer, et~al.
\newblock {Mixture-of-Transformers}: A sparse and scalable architecture for multi-modal foundation models.
\newblock \emph{arXiv preprint arXiv:2411.04996}, 2024.

\bibitem[Liang et~al.(2025)Liang, Li, Yang, Wu, Mao, Nian, Pei, Zhou, Yang, Pang, et~al.]{liang2025discrete}
Zhixuan Liang, Yizhuo Li, Tianshuo Yang, Chengyue Wu, Sitong Mao, Tian Nian, Liuao Pei, Shunbo Zhou, Xiaokang Yang, Jiangmiao Pang, et~al.
\newblock Discrete diffusion vla: Bringing discrete diffusion to action decoding in vision-language-action policies.
\newblock \emph{arXiv preprint arXiv:2508.20072}, 2025.

\bibitem[Lipman et~al.(2022)Lipman, Chen, Ben-Hamu, Nickel, and Le]{lipman2022flow}
Yaron Lipman, Ricky~TQ Chen, Heli Ben-Hamu, Maximilian Nickel, and Matt Le.
\newblock Flow matching for generative modeling.
\newblock \emph{arXiv preprint arXiv:2210.02747}, 2022.

\bibitem[Liu et~al.(2023)Liu, Zhu, Gao, Feng, Liu, Zhu, and Stone]{liu2023libero}
Bo~Liu, Yifeng Zhu, Chongkai Gao, Yihao Feng, Qiang Liu, Yuke Zhu, and Peter Stone.
\newblock {LIBERO}: Benchmarking knowledge transfer for lifelong robot learning.
\newblock \emph{Advances in Neural Information Processing Systems}, 36:\penalty0 44776--44791, 2023.

\bibitem[Liu et~al.(2025)Liu, Chen, An, Liu, Zhang, Gu, Li, Guo, Chen, Liu, et~al.]{liu2025hybridvla}
Jiaming Liu, Hao Chen, Pengju An, Zhuoyang Liu, Renrui Zhang, Chenyang Gu, Xiaoqi Li, Ziyu Guo, Sixiang Chen, Mengzhen Liu, et~al.
\newblock {HybridVLA}: Collaborative diffusion and autoregression in a unified vision-language-action model.
\newblock \emph{arXiv preprint arXiv:2503.10631}, 2025.

\bibitem[Liu et~al.(2024{\natexlab{a}})Liu, Wu, Li, Tan, Chen, Wang, Xu, Su, and Zhu]{liu2024rdt}
Songming Liu, Lingxuan Wu, Bangguo Li, Hengkai Tan, Huayu Chen, Zhengyi Wang, Ke~Xu, Hang Su, and Jun Zhu.
\newblock {RDT-1B}: a diffusion foundation model for bimanual manipulation.
\newblock \emph{arXiv preprint arXiv:2410.07864}, 2024{\natexlab{a}}.

\bibitem[Liu et~al.(2026)Liu, Li, Ma, Wu, Tan, Ouyang, Su, and Zhu]{liu2026rdt2}
Songming Liu, Bangguo Li, Kai Ma, Lingxuan Wu, Hengkai Tan, Xiao Ouyang, Hang Su, and Jun Zhu.
\newblock {RDT2}: Exploring the scaling limit of umi data towards zero-shot cross-embodiment generalization.
\newblock \emph{arXiv preprint arXiv:2602.03310}, 2026.

\bibitem[Liu et~al.(2022)Liu, Gong, and Liu]{liu2022flow}
Xingchao Liu, Chengyue Gong, and Qiang Liu.
\newblock Flow straight and fast: Learning to generate and transfer data with rectified flow.
\newblock \emph{arXiv preprint arXiv:2209.03003}, 2022.

\bibitem[Liu et~al.(2024{\natexlab{b}})Liu, Duan, Zhang, Li, Zhang, Zhao, Yuan, Wang, He, Liu, et~al.]{liu2024mmbench}
Yuan Liu, Haodong Duan, Yuanhan Zhang, Bo~Li, Songyang Zhang, Wangbo Zhao, Yike Yuan, Jiaqi Wang, Conghui He, Ziwei Liu, et~al.
\newblock {MMBench}: Is your multi-modal model an all-around player?
\newblock In \emph{European conference on computer vision}, pages 216--233. Springer, 2024{\natexlab{b}}.

\bibitem[Loshchilov and Hutter(2017)]{loshchilov2017decoupled}
Ilya Loshchilov and Frank Hutter.
\newblock Decoupled weight decay regularization.
\newblock \emph{arXiv preprint arXiv:1711.05101}, 2017.

\bibitem[Lu et~al.(2022)Lu, Mishra, Xia, Qiu, Chang, Zhu, Tafjord, Clark, and Kalyan]{lu2022learn}
Pan Lu, Swaroop Mishra, Tony Xia, Liang Qiu, Kai-Wei Chang, Song-Chun Zhu, Oyvind Tafjord, Peter Clark, and Ashwin Kalyan.
\newblock Learn to explain: Multimodal reasoning via thought chains for science question answering.
\newblock In \emph{The 36th Conference on Neural Information Processing Systems (NeurIPS)}, 2022.

\bibitem[Ma et~al.(2025)Ma, Zhou, Yang, Wang, and Fan]{ma2025running}
Yunchao Ma, Yizhuang Zhou, Yunhuan Yang, Tiancai Wang, and Haoqiang Fan.
\newblock Running vlas at real-time speed.
\newblock \emph{arXiv preprint arXiv:2510.26742}, 2025.

\bibitem[Mees et~al.(2022)Mees, Hermann, Rosete-Beas, and Burgard]{mees2022calvin}
Oier Mees, Lukas Hermann, Erick Rosete-Beas, and Wolfram Burgard.
\newblock {CALVIN}: A benchmark for language-conditioned policy learning for long-horizon robot manipulation tasks.
\newblock \emph{IEEE Robotics and Automation Letters}, 7\penalty0 (3):\penalty0 7327--7334, 2022.

\bibitem[NVIDIA et~al.(2025)NVIDIA, Bjorck, Fernando~Castañeda, Da, Ding, Fan, Fang, Fox, Hu, Huang, Jang, Jiang, Kautz, Kundalia, Lao, Li, Lin, Lin, Liu, Llontop, Magne, Mandlekar, Narayan, Nasiriany, Reed, Tan, Wang, Wang, Wang, Wang, Xiang, Xie, Xu, Xu, Ye, Yu, Zhang, Zhang, Zhao, Zheng, and Zhu]{gr00tn1_2025}
NVIDIA, Johan Bjorck, Nikita~Cherniadev Fernando~Castañeda, Xingye Da, Runyu Ding, Linxi~"Jim" Fan, Yu~Fang, Dieter Fox, Fengyuan Hu, Spencer Huang, Joel Jang, Zhenyu Jiang, Jan Kautz, Kaushil Kundalia, Lawrence Lao, Zhiqi Li, Zongyu Lin, Kevin Lin, Guilin Liu, Edith Llontop, Loic Magne, Ajay Mandlekar, Avnish Narayan, Soroush Nasiriany, Scott Reed, You~Liang Tan, Guanzhi Wang, Zu~Wang, Jing Wang, Qi~Wang, Jiannan Xiang, Yuqi Xie, Yinzhen Xu, Zhenjia Xu, Seonghyeon Ye, Zhiding Yu, Ao~Zhang, Hao Zhang, Yizhou Zhao, Ruijie Zheng, and Yuke Zhu.
\newblock {GR00T} {N1}: An open foundation model for generalist humanoid robots.
\newblock In \emph{ArXiv Preprint}, March 2025.

\bibitem[{Octo Model Team} et~al.(2024){Octo Model Team}, Ghosh, Walke, Pertsch, Black, Mees, Dasari, Hejna, Xu, Luo, Kreiman, Tan, Chen, Sanketi, Vuong, Xiao, Sadigh, Finn, and Levine]{dibya2023octo}
{Octo Model Team}, Dibya Ghosh, Homer Walke, Karl Pertsch, Kevin Black, Oier Mees, Sudeep Dasari, Joey Hejna, Charles Xu, Jianlan Luo, Tobias Kreiman, {You Liang} Tan, Lawrence~Yunliang Chen, Pannag Sanketi, Quan Vuong, Ted Xiao, Dorsa Sadigh, Chelsea Finn, and Sergey Levine.
\newblock Octo: An open-source generalist robot policy.
\newblock In \emph{Proceedings of Robotics: Science and Systems}, Delft, Netherlands, 2024.

\bibitem[Orogat et~al.(2021)Orogat, Liu, and El-Roby]{Orogat2021}
Abdelghny Orogat, Isabelle Liu, and Ahmed El-Roby.
\newblock {CBench}: {T}owards {B}etter {E}valuation of {Q}uestion {A}nswering {O}ver {K}nowledge {G}raphs.
\newblock \emph{Proceedings of the VLDB Endowment (PVLDB)}, 14\penalty0 (8), 2021.

\bibitem[Peebles and Xie(2023)]{peebles2023scalable}
William Peebles and Saining Xie.
\newblock Scalable diffusion models with transformers.
\newblock In \emph{Proceedings of the IEEE/CVF international conference on computer vision}, pages 4195--4205, 2023.

\bibitem[Perez et~al.(2018)Perez, Strub, De~Vries, Dumoulin, and Courville]{perez2018film}
Ethan Perez, Florian Strub, Harm De~Vries, Vincent Dumoulin, and Aaron Courville.
\newblock {FiLM}: Visual reasoning with a general conditioning layer.
\newblock In \emph{Proceedings of the AAAI conference on artificial intelligence}, volume~32, 2018.

\bibitem[Pertsch et~al.(2025)Pertsch, Stachowicz, Ichter, Driess, Nair, Vuong, Mees, Finn, and Levine]{pertsch2025fast}
Karl Pertsch, Kyle Stachowicz, Brian Ichter, Danny Driess, Suraj Nair, Quan Vuong, Oier Mees, Chelsea Finn, and Sergey Levine.
\newblock {FAST}: Efficient action tokenization for vision-language-action models.
\newblock \emph{arXiv preprint arXiv:2501.09747}, 2025.

\bibitem[Qi et~al.(2025)Qi, Wang, Lin, Yi, Ma, Sreenath, and Malik]{qi2025coordinated}
Haozhi Qi, Yen-Jen Wang, Toru Lin, Brent Yi, Yi~Ma, Koushil Sreenath, and Jitendra Malik.
\newblock Coordinated humanoid manipulation with choice policies.
\newblock \emph{arXiv preprint arXiv:2512.25072}, 2025.

\bibitem[Qu et~al.(2025{\natexlab{a}})Qu, Song, Chen, Chen, Gao, Ye, Lv, Shi, Ren, Ruan, et~al.]{qu2025eo}
Delin Qu, Haoming Song, Qizhi Chen, Zhaoqing Chen, Xianqiang Gao, Xinyi Ye, Qi~Lv, Modi Shi, Guanghui Ren, Cheng Ruan, et~al.
\newblock {EO-1}: Interleaved vision-text-action pretraining for general robot control.
\newblock \emph{arXiv preprint arXiv:2508.21112}, 2025{\natexlab{a}}.

\bibitem[Qu et~al.(2025{\natexlab{b}})Qu, Song, Chen, Yao, Ye, Ding, Wang, Gu, Zhao, Wang, et~al.]{qu2025spatialvla}
Delin Qu, Haoming Song, Qizhi Chen, Yuanqi Yao, Xinyi Ye, Yan Ding, Zhigang Wang, JiaYuan Gu, Bin Zhao, Dong Wang, et~al.
\newblock {SpatialVLA}: Exploring spatial representations for visual-language-action model.
\newblock \emph{arXiv preprint arXiv:2501.15830}, 2025{\natexlab{b}}.

\bibitem[Ren et~al.(2024{\natexlab{a}})Ren, Jiang, Liu, Zeng, Liu, Gao, Huang, Ma, Jiang, Chen, Xiong, Zhang, Li, Tang, Yu, and Zhang]{ren2024grounding}
Tianhe Ren, Qing Jiang, Shilong Liu, Zhaoyang Zeng, Wenlong Liu, Han Gao, Hongjie Huang, Zhengyu Ma, Xiaoke Jiang, Yihao Chen, Yuda Xiong, Hao Zhang, Feng Li, Peijun Tang, Kent Yu, and Lei Zhang.
\newblock {Grounding DINO} 1.5: Advance the "edge" of open-set object detection, 2024{\natexlab{a}}.

\bibitem[Ren et~al.(2024{\natexlab{b}})Ren, Liu, Zeng, Lin, Li, Cao, Chen, Huang, Chen, Yan, Zeng, Zhang, Li, Yang, Li, Jiang, and Zhang]{ren2024grounded}
Tianhe Ren, Shilong Liu, Ailing Zeng, Jing Lin, Kunchang Li, He~Cao, Jiayu Chen, Xinyu Huang, Yukang Chen, Feng Yan, Zhaoyang Zeng, Hao Zhang, Feng Li, Jie Yang, Hongyang Li, Qing Jiang, and Lei Zhang.
\newblock {Grounded SAM}: Assembling open-world models for diverse visual tasks, 2024{\natexlab{b}}.

\bibitem[Reuss et~al.(2024{\natexlab{a}})Reuss, Pari, Agrawal, and Lioutikov]{reuss2024efficient}
Moritz Reuss, Jyothish Pari, Pulkit Agrawal, and Rudolf Lioutikov.
\newblock Efficient diffusion transformer policies with mixture of expert denoisers for multitask learning.
\newblock \emph{arXiv preprint arXiv:2412.12953}, 2024{\natexlab{a}}.

\bibitem[Reuss et~al.(2024{\natexlab{b}})Reuss, Ya{\u{g}}murlu, Wenzel, and Lioutikov]{reuss2024multimodal}
Moritz Reuss, {\"O}mer~Erdin{\c{c}} Ya{\u{g}}murlu, Fabian Wenzel, and Rudolf Lioutikov.
\newblock Multimodal diffusion transformer: Learning versatile behavior from multimodal goals.
\newblock \emph{arXiv preprint arXiv:2407.05996}, 2024{\natexlab{b}}.

\bibitem[Reuss et~al.(2025)Reuss, Zhou, R{\"u}hle, Ya{\u{g}}murlu, Otto, and Lioutikov]{reuss2025flower}
Moritz Reuss, Hongyi Zhou, Marcel R{\"u}hle, {\"O}mer~Erdin{\c{c}} Ya{\u{g}}murlu, Fabian Otto, and Rudolf Lioutikov.
\newblock {FLOWER}: Democratizing generalist robot policies with efficient vision-language-action flow policies.
\newblock \emph{arXiv preprint arXiv:2509.04996}, 2025.

\bibitem[Sharma et~al.(2018)Sharma, Ding, Goodman, and Soricut]{sharma2018conceptual}
Piyush Sharma, Nan Ding, Sebastian Goodman, and Radu Soricut.
\newblock {Conceptual Captions}: A cleaned, hypernymed, image alt-text dataset for automatic image captioning.
\newblock In \emph{Proceedings of ACL}, 2018.

\bibitem[Shi et~al.(2025)Shi, Xie, Liu, Sun, Liu, Wang, Zhou, Fan, Zhang, and Huang]{shi2025memoryvla}
Hao Shi, Bin Xie, Yingfei Liu, Lin Sun, Fengrong Liu, Tiancai Wang, Erjin Zhou, Haoqiang Fan, Xiangyu Zhang, and Gao Huang.
\newblock {MemoryVLA}: Perceptual-cognitive memory in vision-language-action models for robotic manipulation.
\newblock \emph{arXiv preprint arXiv:2508.19236}, 2025.

\bibitem[Singh et~al.(2019)Singh, Natarjan, Shah, Jiang, Chen, Parikh, and Rohrbach]{singh2019towards}
Amanpreet Singh, Vivek Natarjan, Meet Shah, Yu~Jiang, Xinlei Chen, Devi Parikh, and Marcus Rohrbach.
\newblock {Towards VQA Models That Can Read}.
\newblock In \emph{Proceedings of the IEEE Conference on Computer Vision and Pattern Recognition}, pages 8317--8326, 2019.

\bibitem[Tang et~al.(2025)Tang, Sun, Zhao, Yang, Lin, Zhang, Hou, Lu, Liu, and Han]{tang2025vlash}
Jiaming Tang, Yufei Sun, Yilong Zhao, Shang Yang, Yujun Lin, Zhuoyang Zhang, James Hou, Yao Lu, Zhijian Liu, and Song Han.
\newblock {VLASH}: Real-time vlas via future-state-aware asynchronous inference.
\newblock \emph{arXiv preprint arXiv:2512.01031}, 2025.

\bibitem[Team et~al.(2025)Team, Abeyruwan, Ainslie, Alayrac, Arenas, Armstrong, Balakrishna, Baruch, Bauza, Blokzijl, et~al.]{team2025gemini}
Gemini~Robotics Team, Saminda Abeyruwan, Joshua Ainslie, Jean-Baptiste Alayrac, Montserrat~Gonzalez Arenas, Travis Armstrong, Ashwin Balakrishna, Robert Baruch, Maria Bauza, Michiel Blokzijl, et~al.
\newblock Gemini {R}obotics: Bringing ai into the physical world.
\newblock \emph{arXiv preprint arXiv:2503.20020}, 2025.

\bibitem[Tian et~al.(2024)Tian, Yang, Zeng, Wang, Lin, Dong, and Pang]{tian2024predictive}
Yang Tian, Sizhe Yang, Jia Zeng, Ping Wang, Dahua Lin, Hao Dong, and Jiangmiao Pang.
\newblock Predictive inverse dynamics models are scalable learners for robotic manipulation.
\newblock \emph{arXiv preprint arXiv:2412.15109}, 2024.

\bibitem[Tong et~al.(2024)Tong, Brown, Wu, Woo, Middepogu, Akula, Yang, Yang, Iyer, Pan, Wang, Fergus, LeCun, and Xie]{tong2024cambrian1fullyopenvisioncentric}
Shengbang Tong, Ellis Brown, Penghao Wu, Sanghyun Woo, Manoj Middepogu, Sai~Charitha Akula, Jihan Yang, Shusheng Yang, Adithya Iyer, Xichen Pan, Ziteng Wang, Rob Fergus, Yann LeCun, and Saining Xie.
\newblock {Cambrian-1}: A fully open, vision-centric exploration of multimodal llms, 2024.
\newblock URL \url{https://arxiv.org/abs/2406.16860}.

\bibitem[Walke et~al.(2023)Walke, Black, Zhao, Vuong, Zheng, Hansen-Estruch, He, Myers, Kim, Du, et~al.]{walke2023bridgedata}
Homer~Rich Walke, Kevin Black, Tony~Z Zhao, Quan Vuong, Chongyi Zheng, Philippe Hansen-Estruch, Andre~Wang He, Vivek Myers, Moo~Jin Kim, Max Du, et~al.
\newblock {BridgeData v2}: A dataset for robot learning at scale.
\newblock In \emph{Conference on Robot Learning}, pages 1723--1736. PMLR, 2023.

\bibitem[Wang et~al.(2025)Wang, Li, Wang, Zhang, Li, Chen, Wang, and Zhang]{wang2025unified}
Yuqi Wang, Xinghang Li, Wenxuan Wang, Junbo Zhang, Yingyan Li, Yuntao Chen, Xinlong Wang, and Zhaoxiang Zhang.
\newblock Unified vision-language-action model.
\newblock \emph{arXiv preprint arXiv:2506.19850}, 2025.

\bibitem[Wen et~al.(2025)Wen, Zhu, Li, Tang, Shen, and Feng]{wen2025dexvla}
Junjie Wen, Yichen Zhu, Jinming Li, Zhibin Tang, Chaomin Shen, and Feifei Feng.
\newblock {DexVLA}: Vision-language model with plug-in diffusion expert for general robot control.
\newblock \emph{arXiv preprint arXiv:2502.05855}, 2025.

\bibitem[Wiedmann et~al.(2025)Wiedmann, Zohar, Mahla, Wang, Li, Frere, von Werra, Gosthipaty, and Marafioti]{wiedmann2025finevisionopendataneed}
Luis Wiedmann, Orr Zohar, Amir Mahla, Xiaohan Wang, Rui Li, Thibaud Frere, Leandro von Werra, Aritra~Roy Gosthipaty, and Andrés Marafioti.
\newblock {FineVision}: Open data is all you need, 2025.
\newblock URL \url{https://arxiv.org/abs/2510.17269}.

\bibitem[Wu et~al.(2023)Wu, Jing, Cheang, Chen, Xu, Li, Liu, Li, and Kong]{wu2023unleashing}
Hongtao Wu, Ya~Jing, Chilam Cheang, Guangzeng Chen, Jiafeng Xu, Xinghang Li, Minghuan Liu, Hang Li, and Tao Kong.
\newblock Unleashing large-scale video generative pre-training for visual robot manipulation.
\newblock \emph{arXiv preprint arXiv:2312.13139}, 2023.

\bibitem[Xiao et~al.(2023)Xiao, Tian, Chen, Han, and Lewis]{xiao2023efficient}
Guangxuan Xiao, Yuandong Tian, Beidi Chen, Song Han, and Mike Lewis.
\newblock Efficient streaming language models with attention sinks.
\newblock \emph{arXiv preprint arXiv:2309.17453}, 2023.

\bibitem[Yang et~al.(2025)Yang, Tan, Wu, Zheng, Peng, Liang, Gu, Cai, Ye, Jang, et~al.]{yang2025magma}
Jianwei Yang, Reuben Tan, Qianhui Wu, Ruijie Zheng, Baolin Peng, Yongyuan Liang, Yu~Gu, Mu~Cai, Seonghyeon Ye, Joel Jang, et~al.
\newblock Magma: A foundation model for multimodal ai agents.
\newblock In \emph{Proceedings of the Computer Vision and Pattern Recognition Conference}, pages 14203--14214, 2025.

\bibitem[Ying et~al.(2025)Ying, Chen, Wang, Jiang, Wang, Yuan, Su, Kong, Yang, and Dong]{ying-etal-2025-seedbench}
Jie Ying, Zihong Chen, Zhefan Wang, Wanli Jiang, Chenyang Wang, Zhonghang Yuan, Haoyang Su, Huanjun Kong, Fan Yang, and Nanqing Dong.
\newblock {S}eed{B}ench: A multi-task benchmark for evaluating large language models in seed science.
\newblock In Wanxiang Che, Joyce Nabende, Ekaterina Shutova, and Mohammad~Taher Pilehvar, editors, \emph{Proceedings of the 63rd Annual Meeting of the Association for Computational Linguistics (Volume 1: Long Papers)}, pages 31395--31449, Vienna, Austria, July 2025. Association for Computational Linguistics.
\newblock ISBN 979-8-89176-251-0.
\newblock URL \url{https://aclanthology.org/2025.acl-long.1516/}.

\bibitem[Yue et~al.(2024)Yue, Ni, Zhang, Zheng, Liu, Zhang, Stevens, Jiang, Ren, Sun, Wei, Yu, Yuan, Sun, Yin, Zheng, Yang, Liu, Huang, Sun, Su, and Chen]{yue2023mmmu}
Xiang Yue, Yuansheng Ni, Kai Zhang, Tianyu Zheng, Ruoqi Liu, Ge~Zhang, Samuel Stevens, Dongfu Jiang, Weiming Ren, Yuxuan Sun, Cong Wei, Botao Yu, Ruibin Yuan, Renliang Sun, Ming Yin, Boyuan Zheng, Zhenzhu Yang, Yibo Liu, Wenhao Huang, Huan Sun, Yu~Su, and Wenhu Chen.
\newblock {MMMU}: A massive multi-discipline multimodal understanding and reasoning benchmark for expert agi.
\newblock In \emph{Proceedings of CVPR}, 2024.

\bibitem[Zhao et~al.(2023)Zhao, Kumar, Levine, and Finn]{zhao2023learning}
Tony~Z Zhao, Vikash Kumar, Sergey Levine, and Chelsea Finn.
\newblock Learning fine-grained bimanual manipulation with low-cost hardware.
\newblock \emph{arXiv preprint arXiv:2304.13705}, 2023.

\bibitem[Zitkovich et~al.(2023)Zitkovich, Yu, Xu, Xu, Xiao, Xia, Wu, Wohlhart, Welker, Wahid, et~al.]{zitkovich2023rt2}
Brianna Zitkovich, Tianhe Yu, Sichun Xu, Peng Xu, Ted Xiao, Fei Xia, Jialin Wu, Paul Wohlhart, Stefan Welker, Ayzaan Wahid, et~al.
\newblock Rt-2: Vision-language-action models transfer web knowledge to robotic control.
\newblock In \emph{Conference on Robot Learning}, pages 2165--2183. PMLR, 2023.

\end{thebibliography}

\clearpage
\appendix
\section*{Appendix}
\addcontentsline{toc}{section}{Appendix}

\section{Case Study for Preservation of Vision-Language Capabilities}
\label{app:vl_competence_demos}

\begin{figure}[!htbp]
    \centering
    \includegraphics[width=\textwidth]{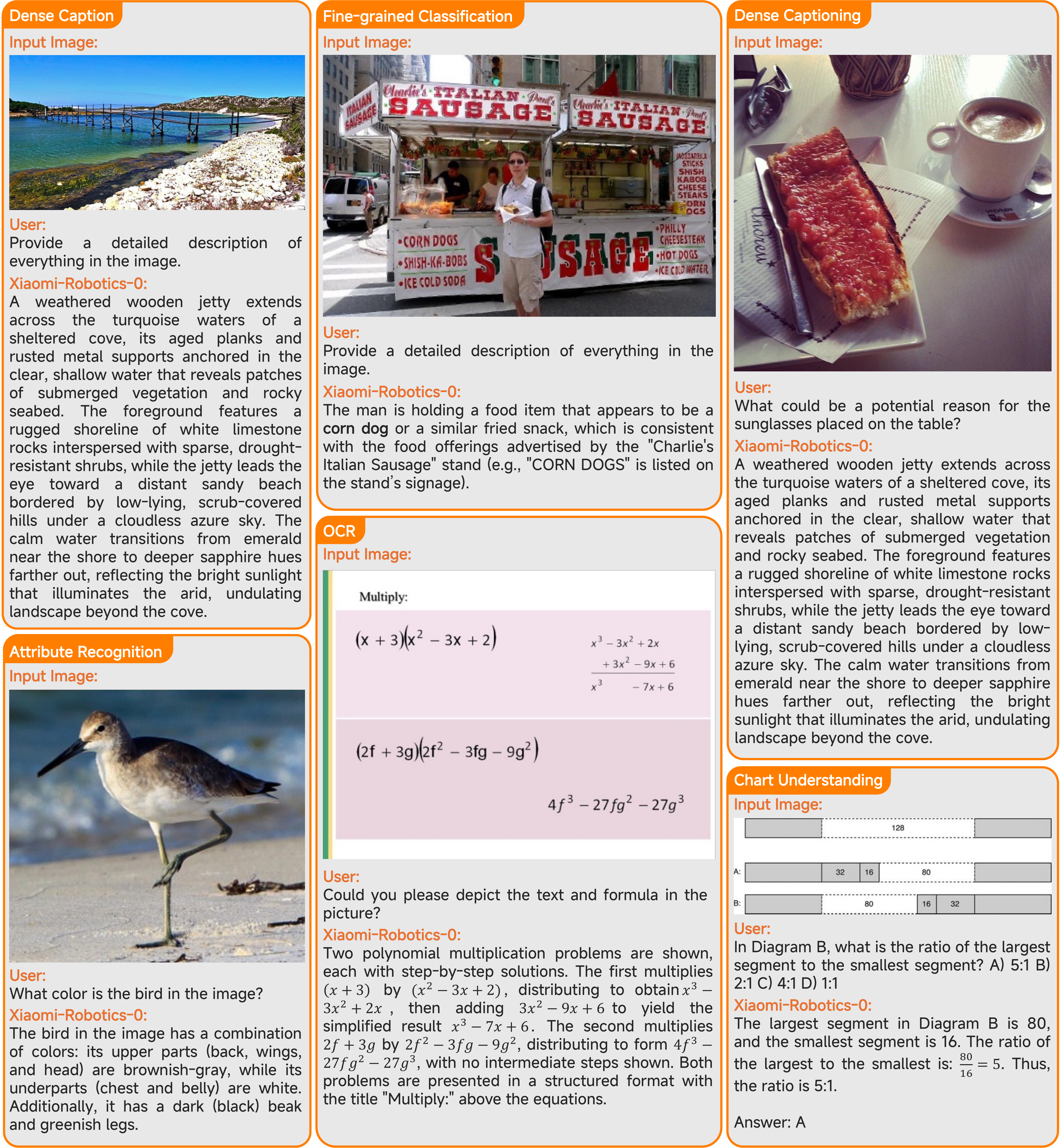}
    \caption{Qualitative results of \Ours{} on vision-language tasks (I).
    }
    \label{fig:vlm_capacity_1}
\end{figure}

\begin{figure}[!htbp]
    \centering
    \includegraphics[width=\textwidth]{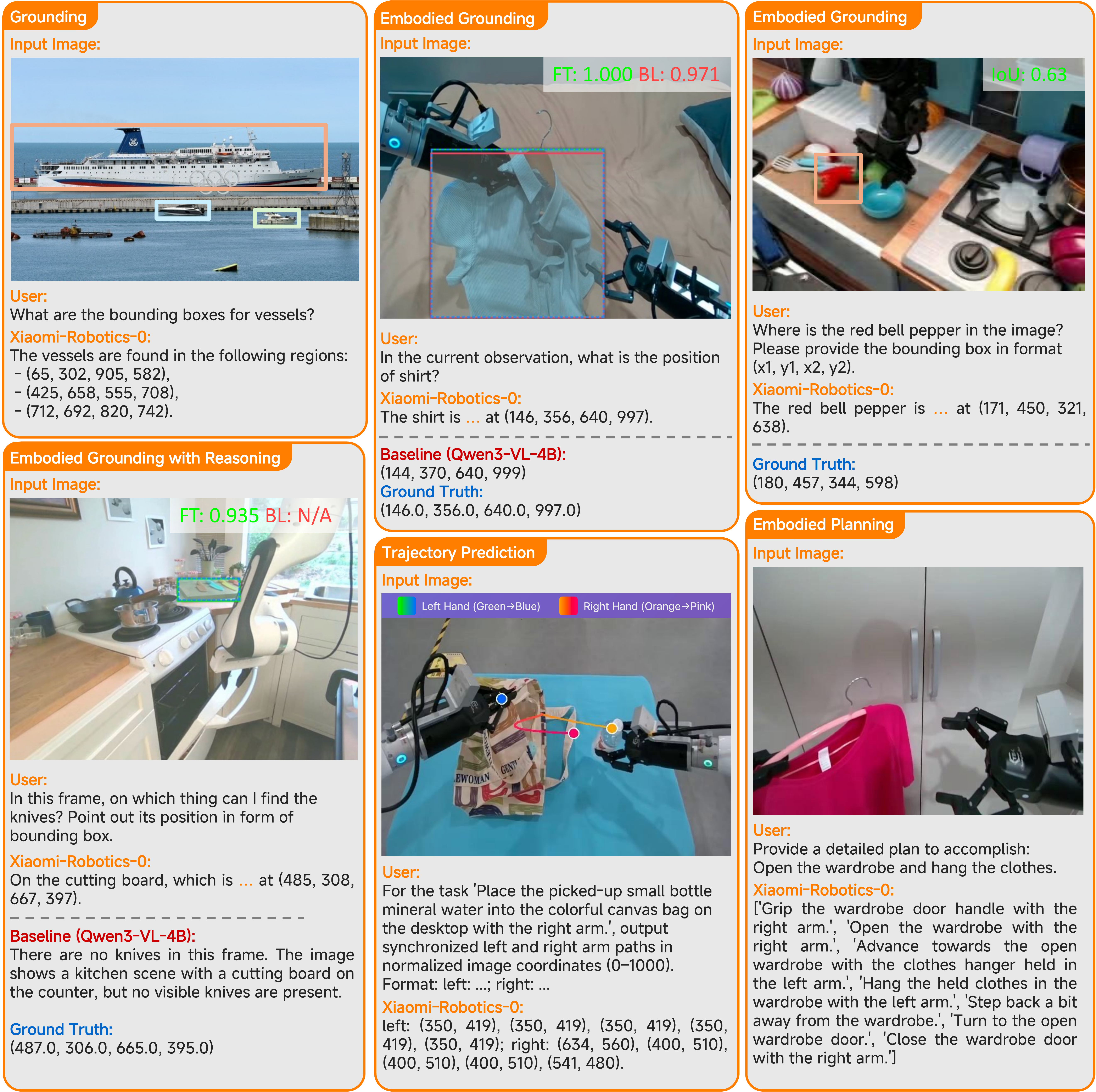}
    \caption{Qualitative results of \Ours{} on vision-language tasks (II). ``\textcolor{orange}{...}'' indicates omitted content for space constraints.
    }
    \label{fig:vlm_capacity_2}
\end{figure}

\begin{figure}[!htbp]
    \centering
    \includegraphics[width=\textwidth]{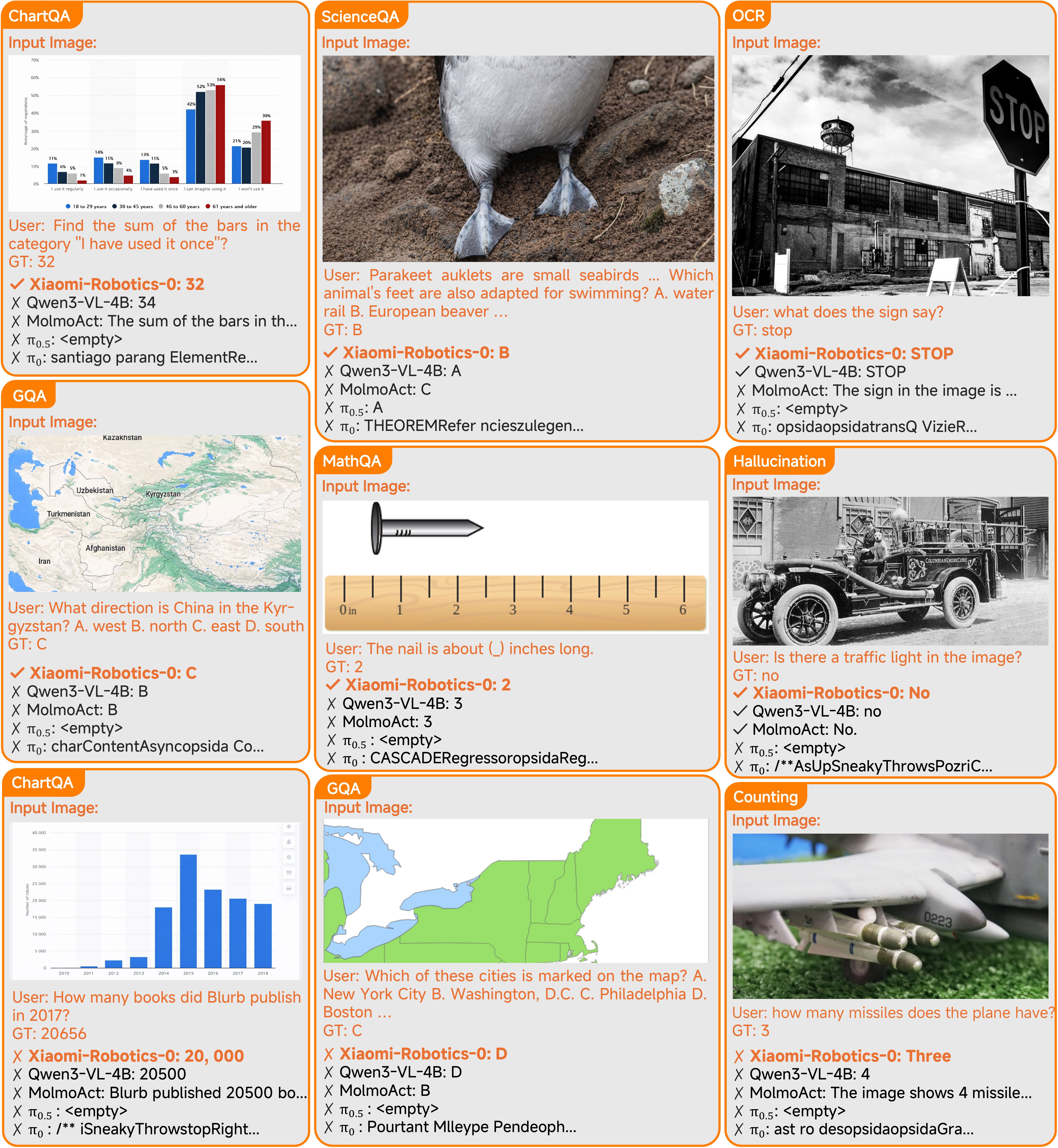}
    \caption{Qualitative comparison of \Ours{} against baseline methods. 
    In particular, the bottom row illustrates challenging failure cases, highlighting limitations in complex numerical reasoning on dense charts and minor format-following errors (\textit{e.g.}, outputting words instead of digits in counting tasks).}
    \label{fig:vlm_capacity_3}
\end{figure}

\newpage

\section{Detailed results on SimplerEnv}
\begin{table}[h]
\centering
\small
\resizebox{\columnwidth}{!}{
\begin{tabular}{l *{5}{>{\centering\arraybackslash}m{2.8cm}}}
\toprule
\textbf{WidowX} & \textbf{Put Spoon on Towel} & \textbf{Put Carrot on Plate} & \textbf{Stack Blocks} & \textbf{Put Eggplant in Basket} & \textbf{Overall} \\ \midrule
RT-1-X~\cite{Neill2023open_x_embodiment}    & 0$\%$    & 4.2$\%$  & 0$\%$    & 0$\%$    & 1.1$\%$  \\
OpenVLA~\cite{kim2024openvla}               & 0$\%$    & 0$\%$    & 0$\%$    & 4.1$\%$  & 1.0$\%$  \\
Octo-Base~\cite{dibya2023octo}              & 12.5$\%$ & 8.3$\%$  & 0$\%$    & 43.1$\%$ & 16.0$\%$ \\
Octo-Small~\cite{dibya2023octo}             & 47.2$\%$ & 9.7$\%$  & 4.2$\%$  & 56.9$\%$ & 29.5$\%$ \\
Magma~\cite{yang2025magma}                  & 37.5$\%$ & 29.2$\%$ & 20.8$\%$ & \underline{91.7}$\%$ & 44.8$\%$ \\
RoboVLMs~\cite{liu2025towards}              & 45.8$\%$ & 20.8$\%$ & 4.2$\%$  & 79.2$\%$ & 37.5$\%$ \\
$\pi_0$~\cite{black2024pi_0}                & \underline{83.8$\%$} & 52.5$\%$ & 52.5$\%$ & 87.9$\%$ & 69.2$\%$ \\
$\pi_0$-FAST~\cite{pertsch2025fast}         & 29.1$\%$ & 21.9$\%$ & 10.8$\%$ & 66.6$\%$ & 32.1$\%$ \\
SpatialVLA~\cite{qu2025spatialvla}          & 16.7$\%$ & 25.0$\%$ & 29.2$\%$ & \textbf{100$\%$} & 42.7$\%$ \\
ThinkAct~\cite{huang2025thinkact}           & 58.3$\%$ & 37.5$\%$ & 8.7$\%$  & 70.8$\%$ & 43.8$\%$ \\
EO-1~\cite{qu2025eo}                        & 63.6$\%$ & \underline{54.5$\%$} & \textbf{81.8}$\%$ & 90.9$\%$ & \underline{72.7$\%$} \\
\rowcolor[HTML]{FFE0CD}
\texttt{\Ours{}} (Ours)   & \textbf{95.8$\%$} & \textbf{62.5$\%$} & \underline{75.0$\%$} & 83.3$\%$ & \textbf{79.2$\%$} \\ \bottomrule
\end{tabular}
}
\caption{Results on the WidowX evaluations of SimplerEnv.}
\label{tab:simplerenv_widowx}
\end{table}

\begin{table}[htbp]
\centering
\small
\resizebox{\columnwidth}{!}{
\begin{tabular}{l *{5}{>{\centering\arraybackslash}m{2.8cm}}}
\toprule
\textbf{Visual Matching} & \textbf{Pick Coke Can} & \textbf{Move Near} & \textbf{Open/Close Drawer} & \textbf{Drawer Apple} & \textbf{Overall} \\ \midrule
Octo-Base~\cite{dibya2023octo}                  & 17.0$\%$ & 4.2$\%$  & 22.7$\%$ & 0$\%$    & 11.0$\%$ \\
OpenVLA~\cite{kim2024openvla}                   & 16.3$\%$ & 46.2$\%$ & 35.6$\%$ & 0$\%$    & 24.5$\%$ \\
RT-1~\cite{brohan2022rt}                        & 85.7$\%$ & 44.2$\%$ & \underline{73.0$\%$} & 6.5$\%$  & 52.4$\%$ \\
RT-1-X~\cite{Neill2023open_x_embodiment}        & 56.7$\%$ & 31.7$\%$ & 59.7$\%$ & 40.7$\%$ & 47.2$\%$ \\
RT-2-X~\cite{Neill2023open_x_embodiment}        & 78.7$\%$ & 77.9$\%$ & 25.0$\%$ & 7.4$\%$  & 47.3$\%$ \\
Magma~\cite{yang2025magma}                      & 75.0$\%$ & 53.0$\%$ & 58.9$\%$ & 8.3$\%$  & 48.8$\%$ \\
RoboVLMs~\cite{liu2025towards}                  & 77.3$\%$ & 61.7$\%$ & 43.5$\%$ & 24.1$\%$ & 51.7$\%$ \\
SpatialVLA~\cite{qu2025spatialvla}              & 86.0$\%$ & 77.9$\%$ & 57.4$\%$ & 0$\%$    & 55.3$\%$ \\
$\pi_0$~\cite{black2024pi_0}                    & 97.9$\%$ & 78.7$\%$ & 62.3$\%$ & 46.6$\%$ & 71.4$\%$ \\
$\pi_0$-FAST~\cite{pertsch2025fast}             & 75.3$\%$ & 67.5$\%$ & 42.9$\%$ & 0$\%$    & 46.4$\%$ \\
ThinkAct~\cite{huang2025thinkact}               & 92.0$\%$ & 72.4$\%$ & 50.0$\%$ & -        & -        \\
MolmoAct~\cite{lee2025molmoact}                 & 77.7$\%$ & 77.1$\%$ & 60.0$\%$ & -        & -        \\
EO-1~\cite{qu2025eo}                            & \underline{98.0$\%$} & \underline{83.8$\%$} & 71.3$\%$ & \underline{52.8$\%$} & \underline{76.5$\%$} \\ 
\rowcolor[HTML]{FFE0CD}
\texttt{\Ours{}} (Ours)                         & \textbf{98.7$\%$} & \textbf{88.8$\%$} & \textbf{79.6$\%$} & \textbf{75.0$\%$} & \textbf{85.5$\%$} \\ \midrule
\textbf{Visual Aggregation} & \textbf{Pick Coke Can} & \textbf{Move Near} &\textbf{Open/Close Drawer} & \textbf{Drawer Apple} & \textbf{Overall} \\ \midrule
Octo-Base~\cite{dibya2023octo}                  & 0.6$\%$  & 3.1$\%$  & 1.1$\%$  & 0$\%$    & 1.2$\%$  \\
OpenVLA~\cite{kim2024openvla}                   & 54.5$\%$ & 47.7$\%$ & 17.7$\%$ & 0.0$\%$  & 30.0$\%$ \\
RT-1~\cite{brohan2022rt}                        & 89.8$\%$ & 50.0$\%$ & 32.3$\%$ & 2.6$\%$  & 43.7$\%$ \\
RT-1-X~\cite{Neill2023open_x_embodiment}        & 49.0$\%$ & 32.3$\%$ & 29.4$\%$ & 10.1$\%$ & 30.2$\%$ \\
RT-2-X~\cite{Neill2023open_x_embodiment}        & 82.3$\%$ & 79.2$\%$ & 35.3$\%$ & 20.6$\%$ & 54.4$\%$ \\
Magma~\cite{yang2025magma}                      & 68.6$\%$ & 78.5$\%$ & 59.0$\%$ & \underline{24.0$\%$} & 57.5$\%$ \\
RoboVLMs~\cite{liu2025towards}                  & 75.6$\%$ & 60.0$\%$ & 10.6$\%$ & 0$\%$    & 36.6$\%$ \\
$\pi_0$~\cite{black2024pi_0}                    & \underline{90.1$\%$} & \underline{80.7$\%$} & 27.6$\%$ & 20.5$\%$ & 54.7$\%$ \\
$\pi_0$-FAST~\cite{pertsch2025fast}             & 77.6$\%$ & 68.2$\%$ & 31.3$\%$ & 0$\%$    & 44.3$\%$ \\
SpatialVLA~\cite{qu2025spatialvla}              & 88.0$\%$ & 72.7$\%$ & 41.8$\%$ & 6.3$\%$  & 52.2$\%$ \\
ThinkAct~\cite{huang2025thinkact}               & 84.0$\%$ & 63.8$\%$ & 47.6$\%$ & -        & -        \\
MolmoAct~\cite{lee2025molmoact}                 & 76.1$\%$ & 61.3$\%$ & \textbf{78.8$\%$} & - & -        \\
EO-1~\cite{qu2025eo}                            & \textbf{91.6$\%$} & \textbf{81.7$\%$} & 55.0$\%$ & 23.8$\%$ & \underline{63.0}$\%$ \\
\rowcolor[HTML]{FFE0CD}
\texttt{\Ours{}} (Ours)                         & 88.2$\%$ & 76.8$\%$ & \underline{67.2$\%$} & \textbf{66.7$\%$} & \textbf{74.7$\%$} \\ \bottomrule
\end{tabular}
}
\caption{Results on the Google Robot evaluations of SimplerEnv.}
\label{tab:simplerenv_google_robot}
\end{table}

\newpage

\section{VLM Benchmark Details}
\label{app:benchmark_details}
\begin{table}[ht]
\centering
\label{tab:benchmark_summary}
\begin{tabularx}{\textwidth}{
  >{\raggedright\arraybackslash}p{3.0cm}
  >{\centering\arraybackslash}p{2.0cm}
  X
}
\toprule
\textbf{Benchmark} & \textbf{Samples} & \textbf{Evaluation Focus} \\
\midrule
\multicolumn{3}{l}{\textbf{Comprehensive Multi-modal Capabilities}} \\
MMBench~\cite{liu2024mmbench} & 4,329 & Comprehensive VLM ability with circular evaluation \\
SEED-Bench~\cite{ying-etal-2025-seedbench} & 14,233 & Fine-grained image-text understanding \& spatial relations \\
MME~\cite{fu2025mme} & 2,374 & Holistic perception and cognition evaluation \\
\midrule
\multicolumn{3}{l}{\textbf{Object Hallucination Evaluation}} \\
POPE~\cite{Li-hallucination-2023} & 9,000 & Object existence polling (Random/Popular/Adversarial) \\
\midrule
\multicolumn{3}{l}{\textbf{Reasoning and Expert Knowledge}} \\
ERQA~\cite{team2025gemini} & 400 & Embodied AI reasoning in physical scenarios \\
ScienceQA~\cite{lu2022learn} & 2,017 & Scientific question answering with chain-of-thought \\
MMMU~\cite{yue2023mmmu} & 900 & Multi-discipline expert-level reasoning \\
\midrule
\multicolumn{3}{l}{\textbf{Fine-grained Visual Perception}} \\
AI2D~\cite{kembhavi2016diagram} & 3,088 & Scientific diagram structure understanding \\
ChartQA~\cite{Orogat2021} & 2,500 & Data visualization interpretation \\
TextVQA~\cite{singh2019towards} & 1,731 & OCR-based reasoning in natural scenes \\
\midrule
\rowcolor[HTML]{FFE0CD} 
\textbf{Total} & \textbf{$\approx$ 40,500} & \textbf{Diverse Generalization Assessment} \\
\bottomrule
\end{tabularx}
\caption{\textbf{Summary of Vision-Language Benchmarks.} We select a diverse set of 10 benchmarks covering comprehensive capabilities, hallucination, reasoning, and fine-grained perception to evaluate the vision-language capabilities of our model.}
\end{table}

\end{document}